\documentclass[10pt,journal,compsoc]{IEEEtran}
\usepackage{amsmath}
\usepackage{amssymb}
\usepackage{array}
\usepackage{graphicx}
\usepackage[usenames, dvipsnames]{color}
\usepackage{rotating}
\usepackage{multirow}
\usepackage{pgfplots}
\usepackage{algorithm}
\usepackage{algorithmicx}
\usepackage[noend]{algpseudocode}
\usepackage{hyperref}

\pgfplotsset{compat=1.5}

\makeatletter
\newcommand{\printfnsymbol}[1]{%
  \textsuperscript{\@fnsymbol{#1}}%
}
\makeatother

\ifCLASSOPTIONcompsoc
  \usepackage[nocompress]{cite}
\else
  \usepackage{cite}
\fi

\ifCLASSINFOpdf
\else
\fi

\begin{document}
%
\title{Mask TextSpotter: An End-to-End Trainable Neural Network for Spotting Text with\\ Arbitrary Shapes}
%
%
%
%

\author{
Minghui Liao\thanks{\printfnsymbol{1} Authors contribute equally.}\printfnsymbol{1},
Pengyuan Lyu\printfnsymbol{1},
Minghang He,
Cong Yao,
Wenhao Wu,
Xiang Bai
\IEEEcompsocitemizethanks{\IEEEcompsocthanksitem 
M. Liao, P. Lyu, M. He, and X. Bai are with the School of Electronic Information and Communications, Huazhong University of Science and Technology, Wuhan, 430074, China. Emails: \{mhliao, minghanghe09, xbai\}@hust.edu.cn, lvpyuan@gmail.com
\IEEEcompsocthanksitem C. Yao and W. Wu are with Megvii (Face++) Inc., Beijing, 100190, China. Emails: yaocong2010@gmail.com, wwh@megvii.com

}
}

\IEEEtitleabstractindextext{%

\begin{abstract}
Unifying text detection and text recognition in an end-to-end training fashion has become a new trend for reading text in the wild, as these two tasks are highly relevant and complementary. In this paper, we investigate the problem of scene text spotting, which aims at simultaneous text detection and recognition in natural images. An end-to-end trainable neural network named as Mask TextSpotter is presented. Different from the previous text spotters that follow the pipeline consisting of a proposal generation network and a sequence-to-sequence recognition network, Mask TextSpotter enjoys a simple and smooth end-to-end learning procedure, in which both detection and recognition can be achieved directly from two-dimensional space via semantic segmentation. Further, a spatial attention module is proposed to enhance the performance and universality.
Benefiting from the proposed two-dimensional representation on both detection and recognition, it easily handles text instances of irregular shapes, for instance, curved text. We evaluate it on four English datasets and one multi-language dataset, achieving consistently superior performance over state-of-the-art methods in both detection and end-to-end text recognition tasks. Moreover, we further investigate the recognition module of our method separately, which significantly outperforms state-of-the-art methods on both regular and irregular text datasets for scene text recognition.
\end{abstract}

}

\maketitle

\IEEEdisplaynontitleabstractindextext

%
\IEEEpeerreviewmaketitle

\IEEEraisesectionheading{\section{Introduction}\label{sec:introduction}}

\IEEEPARstart{R}{eading} text from images/videos is of great values for plentiful real-world applications such as image recognition/retrieval~\cite{bai2018integrating}, geo-location, office automation, and assistance for the blind, as scene text contains quite useful semantics for understanding the world. Scene text reading provides an automatic and rapid way to access the textual information embodied in natural scenes, which is often divided into two sub-problems: scene text detection and scene text recognition. Benefiting from the powerful representation provided by deep neural networks, scene text detection and recognition have achieved significant progress.

Scene text spotting, which aims at concurrently localizing and recognizing text out of natural images, have been extensively studied by previous methods~\cite{wang2011end,neumann2016real,jaderberg2016reading,liao2018textboxes++, Li_2017_ICCV,Busta_2017_ICCV,liu2018fots,he2018end}. The methods following the traditional pipeline \cite{wang2011end,jaderberg2016reading,liao2017textboxes,liao2018textboxes++} treat the procedures of text detection and recognition separately, in which text proposals are first hit by a trained text detector, then fed into a text recognition model. This framework seems simple and straightforward but may lead to sub-optimal performance for both detection and recognition, since the two tasks are relevant and complementary to each other. On the one hand, the recognition results highly rely on the accuracies of detected text proposals. On the other hand, the recognition results are helpful for removing false positive detections. 

\begin{figure}[tbp]
\begin{center}
\includegraphics[width=1.0\linewidth]{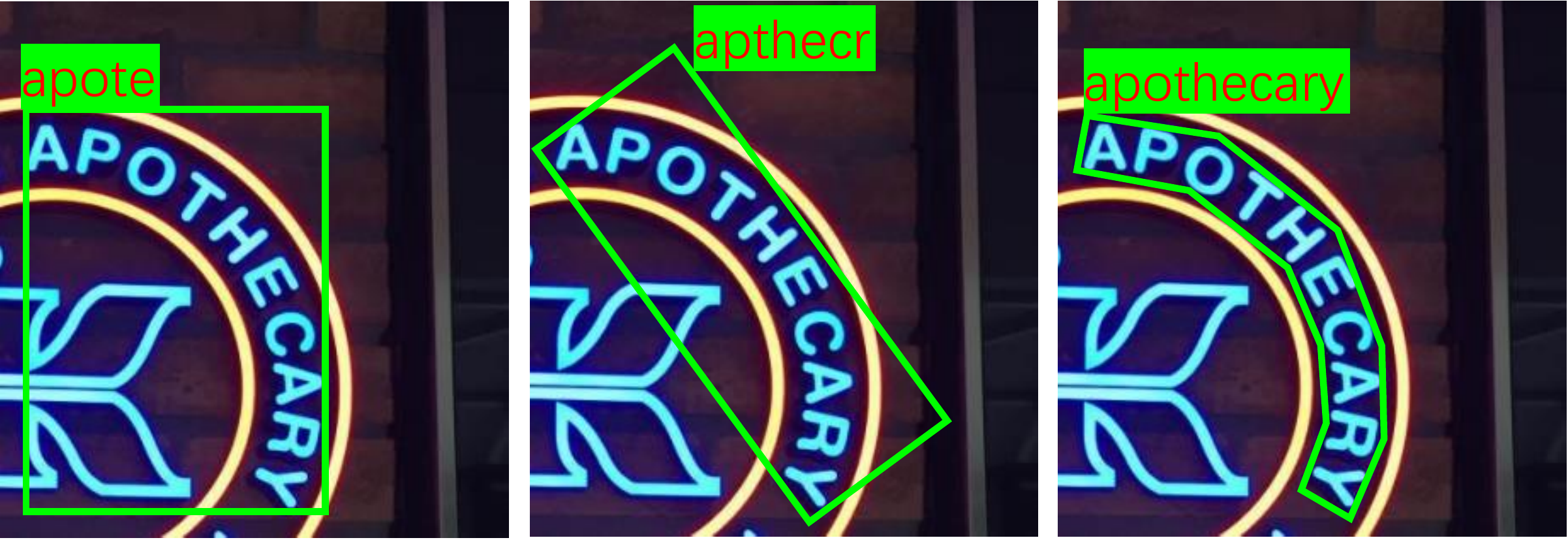}
\end{center}
\caption{Illustrations of different text spotting methods. The left presents horizontal text spotting methods~\cite{liao2017textboxes,Li_2017_ICCV}; The middle indicates oriented text spotting methods~\cite{liao2018textboxes++,Busta_2017_ICCV,liu2018fots}; The right is our proposed method. Green bounding box: detection result; Red text in green background: recognition result.}
\label{fig:introduction}
\end{figure}

Recently, some researchers~\cite{Li_2017_ICCV,Busta_2017_ICCV,liu2018fots,he2018end} start to combine text detection and recognition with an end-to-end trainable network, which consists of two sub-models: a detection network for extracting text instances, and a sequence-to-sequence network for predicting the sequential labels of each text instance. The significant performance improvements for text spotting are achieved by these methods, demonstrating that the detection model and recognition model are complementary in particularly when they are trained in an end-to-end learning fashion. However, these methods suffer from two limitations.

First, they are not completely end-to-end trainable. 
They adopt the curriculum learning paradigm~\cite{he2018end,Li_2017_ICCV,bengio2009curriculum}, 
or apply the alternating training scheme~\cite{Busta_2017_ICCV}, or train the recognition part with the ground truth text regions instead of the predicted proposals~\cite{liu2018fots}.
There are mainly two reasons that stop them from training the models in a smooth, end-to-end fashion. One is that the text recognition part requires accurate locations for training while the predicted locations in the early iterations are usually inaccurate.
The other is that the adopted LSTM~\cite{lstm} or CTC loss~\cite{graves2006connectionist} are more difficult to optimize than general convolutional neural networks (CNNs). 

The second limitation of the above-mentioned methods lies in that these methods only focus on reading the horizontal or oriented text. However, the shapes of text instances in real-world scenarios may vary significantly, from horizontal or oriented, to curved forms. Besides text spotting methods, most state-of-the-art methods for scene text detection~\cite{tian2016detecting,shi2017detecting,zhou2017east,he2017single,he2017deep,liao2018textboxes++} and scene text recognition~\cite{shi2017end,shi2017end} assume that the layout of a text instance is straight, which may fail to handle irregular scene text, for instance, curved text in two dimensional space.

In this paper, we propose an unconventional method for text spotting, named as Mask TextSpotter, which is able to detect and recognize text instances of arbitrary shapes, as shown in Fig.~\ref{fig:introduction}. Here, \textit{arbitrary shapes} mean the various forms of text instances in the real world. Inspired by Mask R-CNN~\cite{he2017mask} that can generate shape masks of objects, our method detects text by segment the instance text regions. Thus our detector is able to detect text of arbitrary shapes. Besides, different from the previous sequence-to-sequence recognition methods~\cite{shi2017end,shi2018aster} which are designed for one-dimensional sequences, we recognize text via semantic segmentation and spatial attention in two-dimensional space, to solve the issues in reading irregular text instances.

In this paper, the major extension over its conference version~\cite{LyuLYWB18} lies in the recognition part of Mask TextSpotter. In \cite{LyuLYWB18}, text recognition is performed by character-level semantic segmentation with the limitation of requiring character location for training and heuristic assumption for character grouping. The proposed improved Mask TextSpotter, which integrates a Spatial Attention Module (SAM) in the recognition part, mitigates the above-mentioned issues greatly.
More specifically, inspired from the recent spatial attention models\cite{donahue2015long,wojna2017attention}, we apply a Spatial Attention Module (SAM) for the recognition part, which can globally predict the label sequence of each word with a spatial attention mechanism. SAM only requires the word-level annotations for training, significantly reducing the need of character-level annotations for training in~\cite{LyuLYWB18}.
In addition, as a global recognition module, SAM is quite complementary to the character segmentation module that locally predicts the character label via the pixel-level classification.

The main contribution in this paper is the proposed Mask TextSpotter, which has several distinct advantages over most state-of-the-art methods on scene text detection, recognition, and spotting.
(1) We tackle the problem of text detection with instance segmentation so that it can detect text of arbitrary shapes.
(2) The recognition model of Mask TextSpotter is more general for handling both regular and irregular text in two-dimensional space, and more effective by simultaneously considering local and global textual information. 
(3) Different from the previous spotting methods that only deal with horizontal or oriented text, the proposed method can spot text of arbitrary shapes, including horizontal, oriented, and curved text. 
(4) Mask TextSpotter is the first framework that is completely end-to-end trainable for text spotting, enjoying a simple, smooth training scheme, so that its detection model and recognition model fully benefit from feature sharing and joint optimization. 
(5) We validate the effectiveness of the proposed method on various English datasets that include horizontal, oriented and curved text, with a single model. Moreover, we verify it on a multi-language dataset to demonstrate the robustness. The results demonstrate that it achieves state-of-the-art performances in both text detection and text spotting on these datasets. More specifically, on ICDAR2015, evaluated at a single scale, our method outperforms the previous top performers by $\mathbf{10.5}$ percents on the end-to-end recognition task with the generic lexicon. We further investigate the standalone recognition model separately out of Mask TextSpotter on the standard datasets for scene text recognition, which significantly outperforms state-of-the-art scene text recognizers.

The rest paper is organized as follows: Sec.~\ref{sec:related-work} reviews the relevant methods. For the methodology, we describe Mask TextSpotter and the standalone recognition model in Sec.~\ref{sec:methodology}. The experiments are discussed and analyzed in Sec.~\ref{sec:experiments}. The conclusion and the future work are summarized in Sec.~\ref{sec:conclusion}.


\section{Related Work}\label{sec:related-work}

\subsection{Scene Text Spotting}
In recent years, scene text detection and recognition have attracted considerable attention from the communities of both computer vision and document analysis. Compared with the methods that only concern one aspect (either scene text detection or text recognition), scene text spotting, or named as end-to-end text recognition that is the most important problem in text information extraction, is relatively less studied by previous methods. 

The earlier methods use handcraft features for scene text spotting. \cite{wang2011end} first detects characters with random ferns and then group them into a word using pictorial structures with a fixed lexicon. Neumann and Matas~\cite{neumann2010method} propose a first lexicon-free end-to-end text recognition system, which performs text detection via MSER. Neumann \emph{et al.}~\cite{neumann2012real,neumann2016real} further improve their system based on Extremal Regions (ER), which significantly improves the accuracy and efficiency of end-to-end text recognition. Yao \emph{et al.}~\cite{yao2014unified} adopt the same features and classification scheme for scene text spotting, which is the first system to cope with both horizontal and multi-oriented scene text.

More recently, deep neural networks have dominated the tasks in scene text detection and recognition. Wang \emph{et al.}~\cite{WangWCN12} attempt to detect characters by a sliding-window-based detector with CNNs and then recognize each character by a character classifier. Bissacco \emph{et al.}~\cite{bissacco2013photoocr} build a reading system, named PhotoOCR, which is able to classify the characters with a DNN model running on HOG features. In \cite{jaderberg2016reading}, Jaderberg \emph{et al.} first generate word proposals using Edge box and aggregated channel features; second, HOG features of these proposals are used for word/non-word classification with a random forest classifier; then a CNN-based bounding box regression method is presented to refine word proposals; finally, a CNN-based word classifier with 90k categories is adopted for word recognition. \cite{jaderberg2016reading} achieves significant improvements over the previous methods for scene text spotting in term of both detection and recognition accuracies. TextBoxes~\cite{liao2017textboxes} simplifies the word detection phase with a single-shot text detector and adopts a sequence-to-sequence text recognizer [63] for word recognition. TextBoxes++~\cite{liao2018textboxes++} further improves \cite{liao2017textboxes} by extending its detection scope from horizontal text to multi-oriented text, and proposes a new scheme for combining detection and recognition stages.   

Unlike all of the above methods that treat the procedures of text detection and text recognition separately, recent  methods make an effort to integrate the detection and recognition models with an end-to-end trainable neural network. Such methods benefit from the complementarity of text detection and recognition. Li \emph{et al.}~\cite{Li_2017_ICCV} combines a single-shot text detector for horizontal text and a sequence-to-sequence text recognizer into a unified network. Meanwhile, \cite{Busta_2017_ICCV} designs a network architecture similar to \cite{Li_2017_ICCV}, but its detection part is more practical for handling both horizontal and multi-oriented text. Then, He \emph{et al.}~\cite{he2018end} and Liu \emph{et al.}~\cite{liu2018fots} follow the similar pipeline of \cite{Li_2017_ICCV,Busta_2017_ICCV}, achieving improvements by incorporating an attentional sequence-to-sequence recognizer~\cite{he2018end} or replacing the detection part with a more robust detection method~\cite{liu2018fots}. In summary, these methods frame end-to-end text recognition as the direct combination of a detector and a sequence-to-sequence recognizer. Instead, our method departs from this strategy by performing text detection and recognition via semantic segmentation and spatial attention from two-dimensional space, resulting in two major advantages over the above methods: 1) It is fully end-to-end trainable, 2) It is able to localize and read text with arbitrary shapes.

\subsection{Scene Text Detection}
Scene text detection plays an important role in the scene text spotting systems.

Deep learning based methods which focus on multi-oriented text have become the mainstream of scene text detection. 
Huang \emph{et al.} \cite{huang2014robust} detect text with CNN induced MSER trees.
Zhang \emph{et al.}  \cite{zhang2016multi} detect multi-oriented scene text by semantic segmentation. 
\cite{tian2016detecting} and \cite{shi2017detecting} propose methods which first detect text segments and then link them into text instances by spatial relationship or link predictions.
Zhou \emph{et al.} \cite{zhou2017east} and He \emph{et al.} \cite{he2017deep} regress text boxes directly from dense segmentation maps. 
Lyu \emph{et al.} \cite{lyu2018multi} propose to detect and group the corner points of the text to generate text boxes. 
Rotation-sensitive regression for oriented scene text detection is proposed by Liao \emph{et al.} \cite{liao2018rotation}.

Recently, detecting text with arbitrary shapes has gradually drawn the attention of researchers due to the application requirements in the real-life scenario. 
Risnumawan \emph{et al.}~\cite{risnumawan2014robust} propose a system for arbitrary text detection based on text symmetry properties.
Based on the symmetric axes of text, Long~\emph{et al.}~\cite{long2018textsnake} propose a flexible representation for text and design a model to regress the radius and orientation from symmetric axes pixels. 

Different from most of the above-mentioned methods, we propose to detect scene text by instance segmentation which can detect text with arbitrary shapes.

\subsection{Scene Text Recognition}

Scene text recognition~\cite{yao2014strokelets,shi2018aster} aims at decoding the detected or cropped image regions into character sequences. 
The previous scene text recognition approaches can be roughly split into three branches: character-based methods, word-based methods, and sequence-to-sequence methods. The character-based recognition methods \cite{bissacco2013photoocr,jaderberg2014deep} mostly first localize individual characters and then recognize and group them into words. 
In \cite{synth90}, Jaderberg \emph{et al.} propose a word-based method which treats text recognition as a common English words (90k) classification problem. 
Sequence-to-sequence methods solve text recognition as a sequence labeling problem. \cite{conf/aaai/HeH0LT16,shi2017end,su2017accurate} use CNN and RNN to model image features and output the recognized sequences with CTC \cite{graves2006connectionist}. In \cite{lee2016recursive}, Lee \emph{et al.} recognize scene text via attention based sequence-to-sequence model. 

Irregular text recognition has recently attracted attention with some methods proposed. In~\cite{shi2018aster}, Shi~\emph{et al.} design a unified network which rectifies the irregular text first with a spatial transform network~\cite{stn} and then recognizes the transformed image via a sequence-to-sequence recognition network. In~\cite{yang2017learning,wojna2017attention}, they propose to recognize irregular text by applying attention mechanism on two-dimensional feature maps. Cheng~\emph{et al.}~\cite{cheng2018aon} propose to encode the input image to four feature sequences of four directions. 

In this paper, we focus on demonstrating the complementarity of the character segmentation module and the spatial attention module, instead of proposing a new spatial attention module. The integration of the character segmentation and the spatial attention not only reduces the need of character level annotations for training, but also makes the model more robust to text shapes.

\subsection{Object Detection and Instance Segmentation}

With the rise of deep learning, object detection\cite{girshick2014rich,fastrcnn,ren2015faster,liu2016ssd,redmon2016you} and semantic/instance segmentation~\cite{long2015fully,dai2016instance,li2017fully,he2017mask} have achieved great development. 
Benefited from those methods, scene text detection and recognition have achieved obvious progress in the past few years. 
Our method is also inspired by those methods. Specifically, our method is inspired by an instance segmentation model Mask R-CNN~\cite{he2017mask}. However, there are key differences between the mask branch of our method and that in Mask R-CNN. Our mask branch can not only segment text regions but also predict character probability maps and text sequences, which means that our method can be used to recognize the instance sequence inside character maps rather than predicting an object mask only.

\section{Methodology}\label{sec:methodology}
\begin{figure*}[htbp]
\begin{center}
\includegraphics[width=0.95\linewidth]{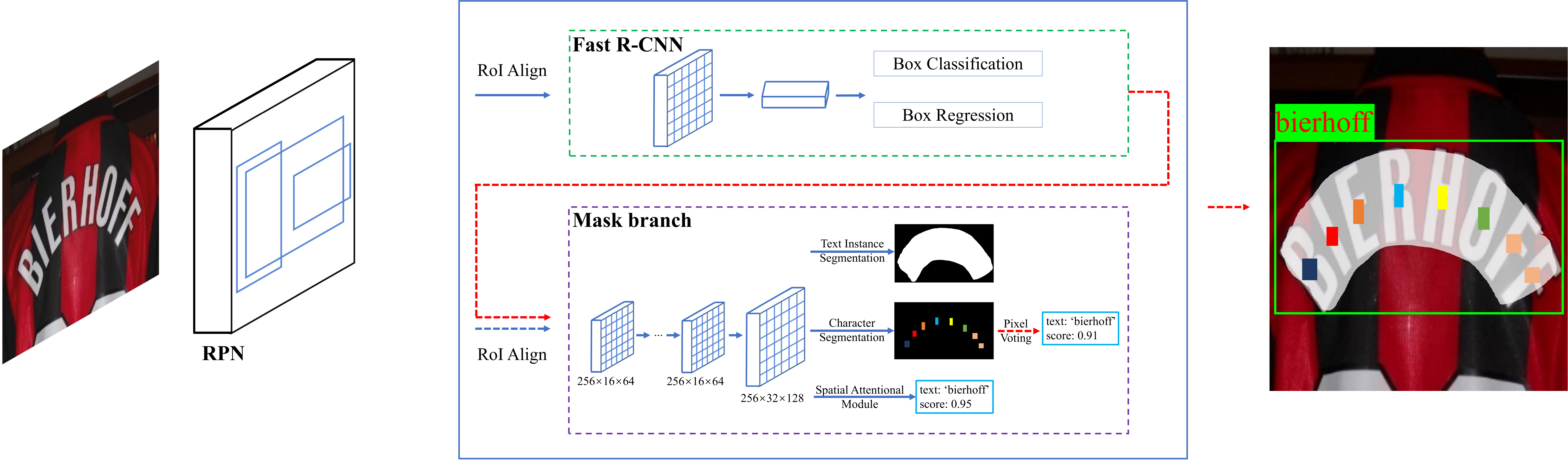}
\end{center}
\caption{Architecture of Mask TextSpotter. The solid arrows mean the data flow both in training and inference period. The dashed arrows in blue and in red indicate the data flow in training stage and inference stage, respectively. The details of the character segmentation and the spatial attentional module are illustrated in Fig.~\ref{fig:architecture}.}
\label{fig:pipeline}
\end{figure*}

\subsection{Architecture}\label{subsec:framework}
Mask TextSpotter is inspired by Mask R-CNN~\cite{he2017mask}. The overall architecture is presented in Fig.~\ref{fig:pipeline}. Functionally, the framework consists of four components: a feature pyramid network (FPN)~\cite{fpn} as backbone, a region proposal network (RPN)~\cite{ren2015faster} for generating text proposals, a Fast R-CNN~\cite{ren2015faster} for bounding boxes regression, a mask branch for text instance segmentation,
character segmentation, and text sequence recognition.

In the training phase, text proposals are first generated by RPN, and then the RoI features of the proposals are fed into the Fast R-CNN branch and the mask branch to generate the accurate text candidate boxes, the text instance segmentation maps, the character segmentation maps, and the text sequence.
\subsubsection{Backbone}\label{subsubsec:backbone}
Text in natural images are various in sizes. In order to build high-level semantic feature maps at all scales, we apply a feature pyramid structure~\cite{fpn} backbone with ResNet-50~\cite{resnet}.
FPN uses a top-down architecture to fuse the feature of different resolutions from a single-scale input, which improves accuracy with marginal cost.
\subsubsection{RPN}
RPN is used to generate text proposals for the subsequent Fast R-CNN and mask branch. Following~\cite{fpn}, we assign anchors on different stages depending on the anchor size. Specifically, the area of the anchors are set to \{$32^2, 64^2, 128^2, 256^2, 512^2$\} pixels on five stages \{$P_2, P_3, P_4, P_5, P_6$\} respectively. Different aspect ratios \{$0.5,1,2$\} are also adopted in each stages as in~\cite{ren2015faster}. In this way, the RPN can handle the text of various sizes and aspect ratios. RoI Align~\cite{he2017mask} is adapted to extract the region features of the proposals. Compared to RoI Pooling~\cite{fastrcnn}, RoI Align preserves more accurate location information, which is quite beneficial to the segmentation task in the mask branch. 

\subsubsection{Fast R-CNN}
The Fast R-CNN branch includes a classification task and a regression task. The main function of this branch is to provide accurate bounding boxes for detection. The inputs of Fast R-CNN are in $7 \times 7$ resolution, which are generated by RoI Align from the proposals produced by RPN.

\subsubsection{Mask Branch}
The mask branch plays the role of detecting and recognizing the text of arbitrary shapes. There are three tasks in the mask branch, including a text instance segmentation task, a character segmentation task and a text sequence recognition task. We will describe them in detail in Sec.~\ref{subsec:seg} and Sec.~\ref{subsec:seq}.

\subsection{Text Instance and Character Segmentation}\label{subsec:seg}
As shown in Fig.~\ref{fig:pipeline}, giving an input RoI feature, whose size is fixed to $16 \times 64$, through four convolutional layers with $3 \times 3$ filters and a de-convolutional layer with $2 \times 2$ filters and strides, the features are fed into two modules. 
A 1-channel text instance map is generated by a convolutional layer, which can give accurate localization of a text region, regardless of the shape of the text instance. In character segmentation module, the character segmentation maps are generated from the shared feature maps directly. The output character maps are of shape $N_s \times 32 \times 128$, where $N_s$ represents the number of classes, which is set to $37$, including 36 for alphanumeric characters and $1$ for the background.

\subsection{Spatial Attentional Module (SAM)}\label{subsec:seq}
There are some limitations in the character segmentation. First, the character segmentation needs the character-level annotations to supervise the training. Second, a specially designed post-processing algorithm is required to yield text sequence from segmentation maps. Third, the order of the characters cannot be obtained from the segmentation maps. Though, by means of some rules, the characters can be grouped to text sequence, the generality is still limited.

\begin{figure*}[!ht]
\centering
\includegraphics[width=0.95\linewidth]{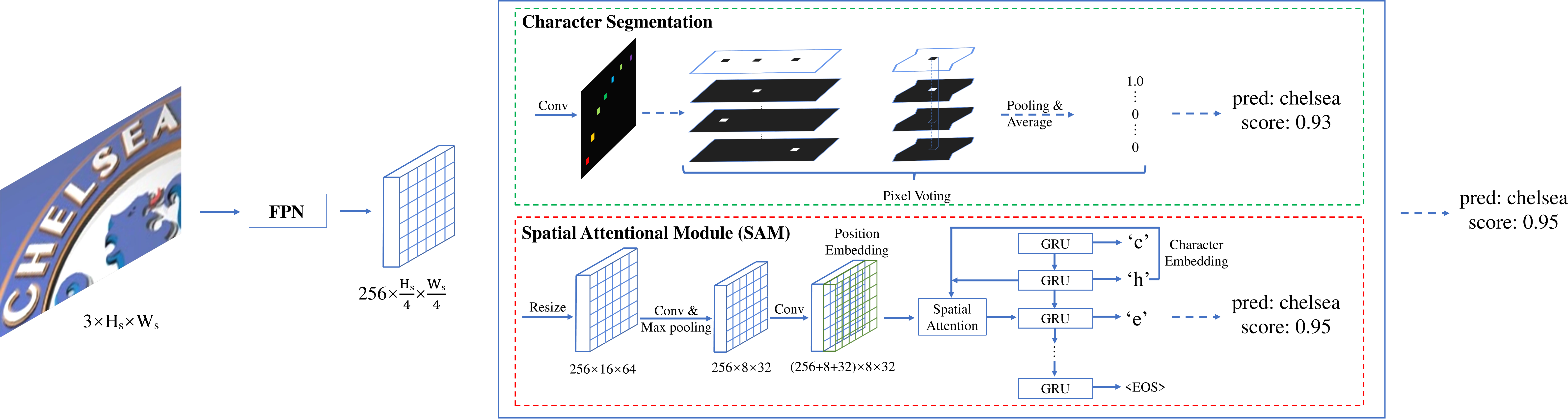}
\caption{Architecture of the standalone recognition model. We use a feature-pyramid structure with ResNet-50. Note that both the two modules can provide the recognition results along with their confidence score, we select the final recognition result with a higher confidence score dynamically. The solid arrows mean the steps in both the training and the inference period; the dashed arrows indicate the steps only in the inference period.}
\label{fig:architecture}
\end{figure*}

To overcome these limitations, inspired from the recent spatial attention models\cite{donahue2015long,wojna2017attention}, we introduce a spatial attentional module (SAM) to decode the text sequence from the feature map in an end-to-end manner. Different from the previous method~\cite{shi2018aster}, which first encode the feature map into a one-dimensional feature sequence and then decode it, SAM directly decodes the two-dimensional feature map for better representation of various shapes. 
The whole pipeline of SAM is illustrated in Fig.~\ref{fig:architecture}. First, a given feature map, which can be the RoI feature in Mask TextSpotter or the feature map from the backbone in the standalone recognition model, is resized to a fixed shape by bilinear interpolation. 
Then, a convolution layer, a max pooling layer, and a convolution layer are performed in order. Finally, the spatial attention with RNNs is applied to produce the text sequence. 

\subsubsection{Position Embedding}\label{subsubsec:position_embedding}
Inspired by \cite{GehringAGYD17,wojna2017attention}, the position embedding is adopted because the transformation operators in SAM are not position-sensitive. We apply a similar position embedding mechanism proposed in \cite{wojna2017attention}.
As shown in Fig.~\ref{fig:architecture}, the position embedding is applied after the last convolution layer. The position embedding feature map $F_{pe}$ is of shape $(W_p+H_p, H_p, W_p)$, where $H_p$, $W_p$ are set to $8$ and $32$ respectively. The position embedding feature map is calculated as below:
\begin{align}
    &F_{pe}^x (i, j, :) = onehot(i,W_p)\\
    &F_{pe}^y (i, j, :) = onehot(j,H_p)\\
    &F_{pe} = Concat(F_{pe}^x, F_{pe}^y)
\end{align}
where $onehot(i,K)$ means a vector $V$ of length $K$, in which the value of the element with the index $i$ is set to $1$ while rest of the values are set to $0$.
We cascade the position embedding feature map with the original input feature map. The cascaded feature map $F$ is of shape $(C+H_p+W_p, H_p, W_p)$, where $C$ is the number of the channel of the original input feature map which is set to $256$.

\subsubsection{Spatial Attention with RNNs}\label{subsubsec:spatial_attention}
Our attention mechanism is based on \cite{bahdanau2014neural}. However, we extend it to a more general form, which learns attentional weights in two-dimensional space. Assume that it works iteratively for $T$ steps, which predicts a sequence of character classes $y=(y_1, ..., y_T)$. At step $t$, there are three inputs: (1) the input feature map $F$ mentioned in Sec.~\ref{subsubsec:position_embedding}; (2) the last hidden state $s_{t-1}$; (3) the last predicted character class $y_{t-1}$. 

First, we expand $s_{t-1}$ from a vector to a feature map $S_{t-1}$ which is of shape $(V,H_p,W_p)$ by copying, where $V$ is the hidden size of the RNN, which is set to $256$.
\begin{equation}
    S_{t-1} = expand\_dim(s_{t-1}, H_p, W_p)
\end{equation}

Then, we calculate the attention weights $\alpha_t$ as follows:
\begin{align}
    e_t &= W_t \times \tanh({W_s \times S_{t-1} + W_f \times F + b})\\
    \alpha_t(i, j) &= exp(e_t(i, j)) / \sum_{i'=1}^{H_p}\sum_{j'=1}^{W_p}exp(e_t(i', j'))
\end{align}
where $e_t$ and $\alpha_t$ is of shape $(H_p, W_p)$. $W_t$, $W_s$, $W_f$ and $b$ are trainable weights and biases.

Next, we can acquire the glimpse $g_t$ of step $t$ by applying the attention weights to the original feature map $F$.
\begin{equation}
    g_t = \sum_{i=1}^{H_p}\sum_{j=1}^{W_p} \alpha_t(i, j) \times F(i,j)
\end{equation}

The RNN input $r_t$ is cascaded by the glimpse $g_t$ and a character embedding of the last predicted character class $y_{t-1}$.
\begin{align}
    f(y_{t-1}) &= W_y \times onehot(y_{t-1}, N_c) + b_y\\
    r_t &= concat(g_t, f(y_{t-1}))
\end{align}
where $W_y$ and $b_y$ are trainable weights and bias of the linear transformation. $N_c$ is the number of classes in the sequence decoder, which is set to $37$, including $36$ classes for alphanumeric characters, $1$ classes for end-of-sequence symbol (EOS).

We feed the RNN input $r_t$ and the last hidden state of RNN $s_{t-1}$ into the RNN cell.
\begin{equation}
    (x_t, s_t) = rnn(s_{t-1}, r_t)
\end{equation}

Finally, the conditional probability at step $t$ is calculated by a linear transformation and a softmax function.
\begin{align}
    p(y_t) &= softmax(W_o \times x_t + b_o) \label{eq:prob}\\
    y_t &\sim p(y_t)
\end{align}

\subsection{Standalone Recognition Model}\label{subsec:stand_alone}
To better verify the superiority of our recognition part, we also build a standalone recognition model.

The overview of the architecture is described in Fig.~\ref{fig:architecture}. 
We use a feature-pyramid structure which is inherited from Feature Pyramid Network (FPN)~\cite{fpn} based on ResNet-50~\cite{resnet}. A Pyramid Pooling Module (PPM)~\cite{zhao2017pyramid} is applied in the last stage of ResNet-50 to enlarge the receptive field. 
Besides, different from the original FPN~\cite{fpn}, we do not down-sample the last two stages while keep their resolution by using dilated convolution~\cite{dilatedconv}. The shared feature map for both the character segmentation module and SAM is generated by up-sampling and concatenating the feature maps from the feature-pyramid structure.

The standalone recognition model consists of two recognition modules. One
is a character segmentation module which predicts the characters at pixel level. A pixel voting algorithm, which groups and arranges the pixels to form the final text sequence result, can be applied to it. Another module is SAM, which predicts text sequence both in two-dimensional perspective and an end-to-end manner.

\subsection{Label Generation}
\begin{figure}[htbp]
\begin{center}
\includegraphics[width=1.0\linewidth]{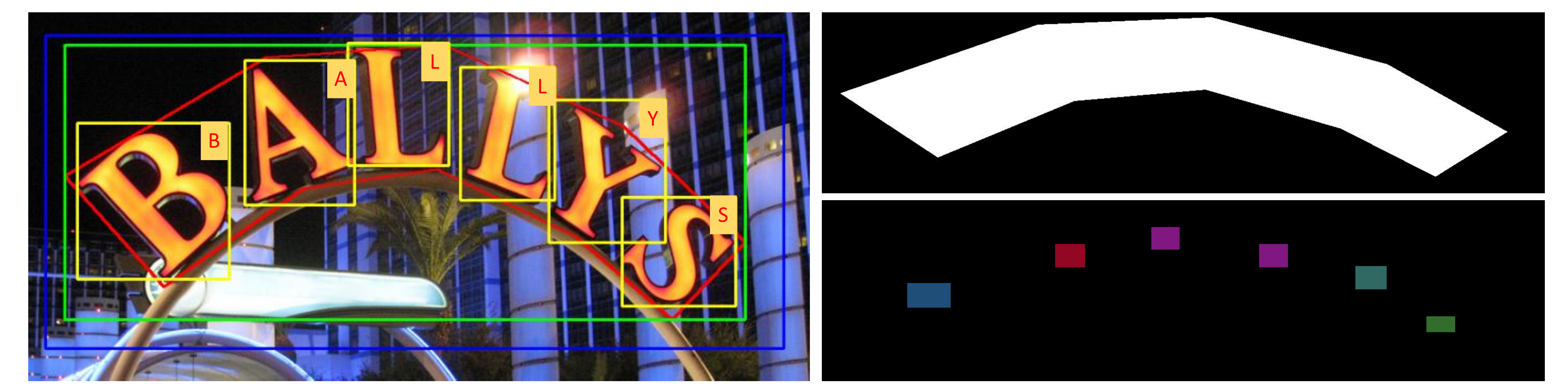}
\end{center}
\caption{Label generation of the text instance segmentation and the character segmentation. Left: the blue box is a proposal yielded by RPN, the red polygon and yellow boxes are ground truth polygon and character boxes, the green box is the horizontal rectangle which covers the polygon with minimal area. Right: the text instance map (top) and the character map (bottom).}
\label{fig:label_generation}
\end{figure}

The label generation of the text instance segmentation and the character segmentation is illustrated in Fig.~\ref{fig:label_generation}.
For a training sample with the input image $I$ and the corresponding ground truth, we generate targets for RPN, Fast R-CNN and the mask branch. Generally, the ground truth contains $P=\{p_{1}, p_{2}...p_{m} \}$ and $C=\{c_{1}=(cc_{1},cl_{1}),c_{2}=(cc_{2},cl_{2}), ... , c_{n}=(cc_{n},cl_{n})\}$, where $p_{i}$ is a polygon which represents the localization of a text region, $cc_{j}$ and $cl_{j}$ are the category and location of a character respectively. Note that, $C$ is not necessary for all training samples. 

We first transform the polygons into horizontal rectangles which cover the polygons with minimal areas. And then we generate targets for RPN and Fast R-CNN following \cite{fastrcnn,ren2015faster,fpn}. There are two types of target maps to be generated for the mask branch with the ground truth $P$, $C$ (may not exist) as well as the proposals yielded by RPN: a map for text instance segmentation and a character map for character semantic segmentation. 
Given a positive proposal $r$, we first use the matching mechanism of \cite{fastrcnn,ren2015faster,fpn} to obtain the best matched horizontal rectangle. The corresponding polygon as well as characters (if any) can be obtained further. Next, the matched polygon and character boxes are shifted and resized to align the proposal and the target map of $H\times W$ as the following formulas:
\begin{equation}
B_{x}=(B_{x_0}-min(r_{x})) \times W / (max(r_{x})-min(r_{x}))
\end{equation}
\begin{equation}
B_{y}=(B_{y_0}-min(r_{y})) \times H / (max(r_{y})-min(r_{y}))
\end{equation}
where $(B_{x},B_{y})$ and $(B_{x_0},B_{y_0})$ are the updated and original vertexes of the polygon and all character boxes; $(r_{x}, r_{y})$ are the vertexes of the proposal $r$.

After that, the target text instance map can be generated by drawing the normalized polygon on a zero-initialized mask and filling the polygon region with the value $1$. 

As for the character map generation, we first shrink all character bounding boxes by fixing their center points and shortening the sides to the fourth of the original sides. Then, the values of the pixels in the shrunk character bounding boxes are set to their corresponding category indices and those outside the shrunk character bounding boxes are set to $0$. If there is no character bounding boxes annotations, all values are set to $-1$, which will be ignored when training.

For SAM, word-level labels are provided, which is expressed as a sequence of character category indices, without localization information.

\subsection{Optimization}
As discussed above, our model includes multiple tasks. We therefor define a multi-task loss function:
\begin{equation}
L = L_{rpn} + \alpha_1 L_{rcnn} + \alpha_2 L_{mask},
\end{equation}
where $L_{rpn}$ and $L_{rcnn}$ are the loss functions of RPN and Fast R-CNN, which are identical as these in~\cite{ren2015faster} and~\cite{fastrcnn}. 
The mask loss $L_{mask}$ consists of a text instance segmentation loss $L_{ins}$, a character segmentation loss $L_{seg}$ and a sequence recognition loss $L_{seq}$:
\begin{equation}
L_{mask} = L_{ins} + \beta_1 L_{seg} + \beta_2 L_{seq},
\end{equation}

Specifically, $L_{ins}$ is an average binary cross-entropy loss; $L_{seg}$ is a weighted spatial soft-max loss which we formulate it as follows:
\begin{equation}
L_{seg} = -\frac{1}{N}\sum_{n=1}^{N}W_n\sum_{c=0}^{N_c-1} Y_{n,c} log(\frac{e^{X_{n,c}}}{\sum_{k=0}^{N_c-1} e^{X_{n,k}}}),
\end{equation}
where $N_c$ is the number of classes, $N$ is the number of pixels in each map and $Y$ is the corresponding ground truth of the output maps $X$. We use weight $W$ to balance the loss value of the positives (character classes) and the background class. Let the number of the background pixels be $N_{neg}$, and the background class index is $0$, the weights can be calculated as:
\begin{equation}
  W_i = 
  \begin{cases}
    1& \text{if } Y_{i,0}=1, \\
    N_{neg} / (N - N_{neg})& \text{otherwise}
  \end{cases}
\end{equation}
The $L_{seq}$ is calculated as follows:
\begin{equation}
L_{seq} = -\sum_{t=1}^{T} \log(p(y_t)),
\end{equation}
where $p(y_t)$ is described in Eq.~\ref{eq:prob}. $T$ is the length of the sequence labels.

In this work, the $\alpha_1$, $\alpha_2$, $\beta_1$, are empirically set to $1.0$, and $\beta_2$ is set to $0.2$.

\subsection{Inference}
\subsubsection{Overview}
Different from the training process where the input RoIs of mask branch come from RPN, in the inference phase, we use the outputs of Fast R-CNN as proposals to generate the predicted text instance maps, character maps, and the text sequence, since the Fast R-CNN outputs are more accurate. 

More specifically, the processes of inference are as follows: first, inputting a test image, we obtain the outputs of Fast R-CNN as \cite{ren2015faster} and filter out the redundant candidate boxes by NMS; and then, the kept proposals are fed into the mask branch to generate the text instance maps, the character maps, and the text sequences; finally the predicted polygons can be obtained directly by calculating the contours of text regions on text instance maps. Besides, the text sequence can be obtained by decoding character segmentation maps and the outputs of SAM, the details are detailed in Sec.~\ref{subsubsec:rec_inference}.

In addition, when performing inference with a lexicon, a weighted edit distance algorithm is proposed to find the best matching word.
\subsubsection{Decoding}\label{subsubsec:rec_inference}

\begin{algorithm}
\scriptsize
\caption{Pixel Voting} 
{\bf Input:} 
Background map $B$, Character maps $C$ 
\begin{algorithmic}[1]
\State Generating connected regions $R$ on the binarized background map
\State $S\leftarrow \varnothing$
\For{$r$ in $R$} 
    \State $scores \leftarrow \varnothing $
    \For{$c$ in $C$}
        \State mean = Average(c[r])
        \State $scores \leftarrow$ scores + {mean}
    \EndFor
    
    \State $S \leftarrow$ S + {Argmax(scores)}
\EndFor
\State \Return $S$
\end{algorithmic}
\label{pixel_voting}
\end{algorithm}

\begin{figure*}[htbp]
\begin{center}
\includegraphics[width=0.95\linewidth]{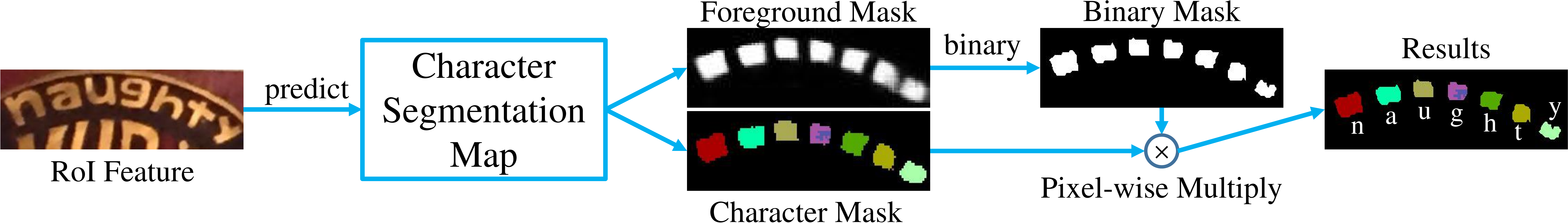}
\end{center}
\caption{Illustration of the pixel voting algorithm. We use the original image crop represents the corresponding RoI feature for better visualization.}
\label{fig:pixelvoting}
\end{figure*}

\noindent\textbf{Character Segmentation}\label{subsubsec:seg_inference}
We decode the predicted character maps into character sequences by our proposed ~\emph{pixel voting} algorithm. 
We first binarize the background map, where the values are from $0$ to $1$, with a threshold of $0.75$.
Then we obtain all character regions according to connected regions in the binarized map. We calculate the mean values of each region for all character maps. The values can be seen as the character-class probability of the region, which is viewed as the confidence scores of characters. The character class with the largest mean value will be assigned to the region. The specific processes are shown in Algorithm~\ref{pixel_voting}. After that, we group all the characters from left to right according to the writing habit of English. A visualized pipeline of the pixel voting algorithm is shown in Fig.~\ref{fig:pixelvoting}.

\noindent\textbf{SAM}\label{subsubsec:sam_inference}
There are two decoding schemes for SAM to produce the final text sequence. One is a greedy decoding strategy which selecting the class with the highest probability at each step. Another is a beam search scheme which maintains top $k$ probabilities at each step. Following previous methods~\cite{shi2017end,shi2018aster}, we adopt the beam search scheme and set $k$ as $6$.

\noindent Since there are two recognition results, we can combine them for better accuracy. The confidence score of the character segmentation module is the mean of all character confidence scores which are mentioned in Algorithm~\ref{pixel_voting} and the confidence score of SAM is the mean of character probabilities in Eq.~\ref{eq:prob}. Naturally, we select the recognition result with a higher confidence score dynamically.

\subsubsection{Weighted Edit Distance} Edit distance can be used to find the best-matched word of a predicted sequence with a given lexicon. However, there may be multiple words matched with the minimal edit distance at the same time, and the algorithm can not decide which one is the best. The main reason for the above-mentioned issue is that all operations (delete, insert, replace) in the original edit distance algorithm have the same costs, which does not make sense actually. 
\begin{figure}[htbp]
\begin{center}
\includegraphics[width=1.0\linewidth]{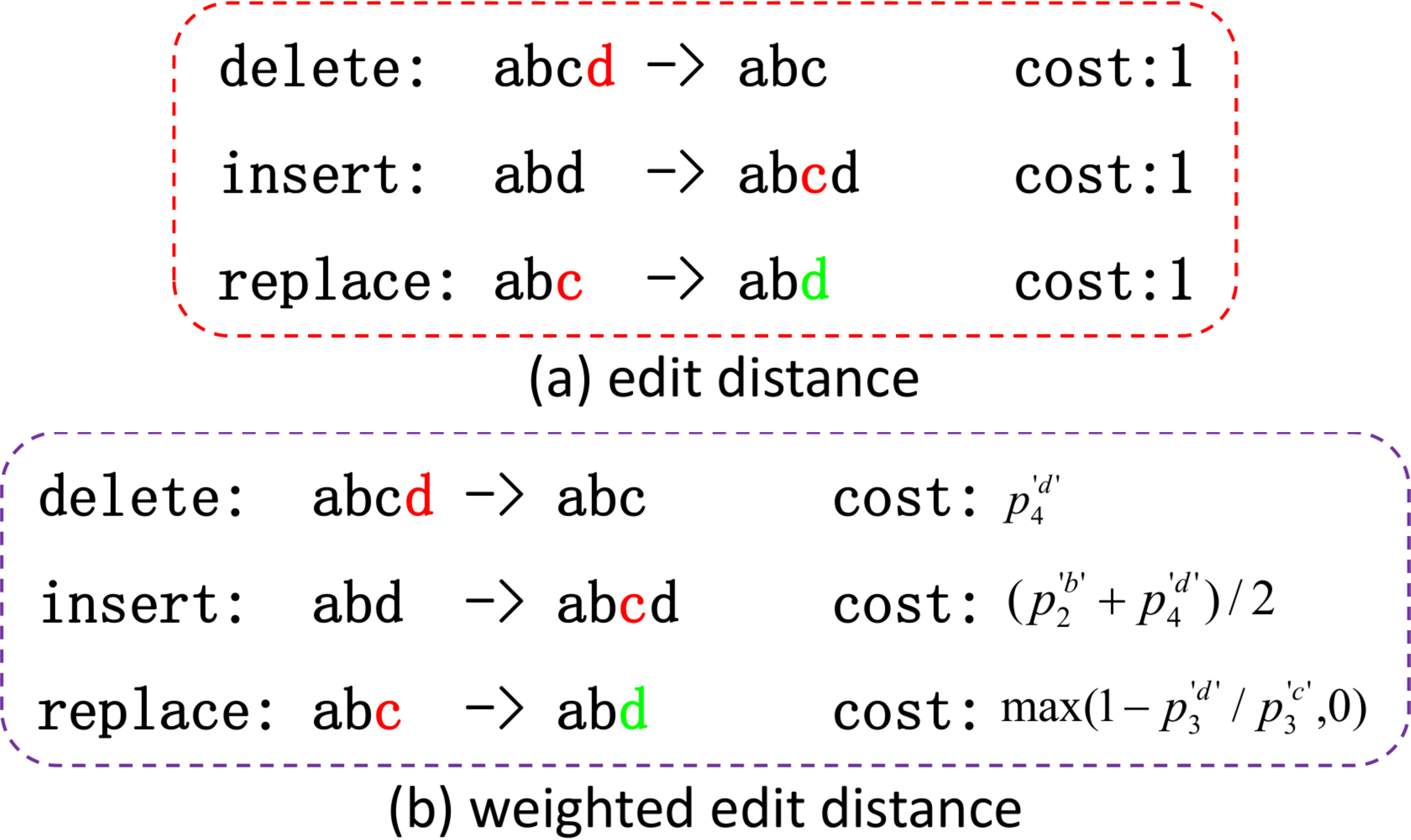}
\end{center}
\caption{Illustration of the edit distance and our proposed weighted edit distance. The red characters are the characters will be deleted, inserted and replaced. Green characters mean the candidate characters. $p_{index}^c$ is the character probability, $index$ is the character index and $c$ is the current character.}
\label{fig:wed}
\end{figure}

Inspired by~\cite{yao2014unified}, we propose a weighted edit distance algorithm. As shown in Fig.~\ref{fig:wed}, different from edit distance, which assign the same cost for different operations, the costs of our proposed weighted edit distance depend on the character probability $p_{index}^c$ which is yielded by the pixel voting or the beam search decoding. Mathematically, the weighted edit distance between two strings $a$ and $b$, whose length are $|a|$ and $|b|$ respectively, can be described as $D_{|a|,|b|}$, which is calculated as following:
\begin{equation}
{D_{i,j}={\begin{cases}\max(i,j)&{\text{ if }}\min(i,j)=0,\\\min {\begin{cases} D_{i-1,j}+C_d\\ D_{i,j-1}+C_i\\ D_{i-1,j-1}+C_r \times 1_{i,j}\end{cases}}&{\text{ otherwise.}}\end{cases}}}
\end{equation}
\normalsize
where $1_{i,j}$ is the indicator function equal to 0 when $a_{i}=b_{j}$ and equal to 1 otherwise; $D_{i,j}$ is the distance between the first $i$ characters of $a$ and the first $j$ characters of $b$; $C_d$, $C_i$, and $C_r$ are the deletion, insert, and replace cost respectively. In contrast, these costs are set to $1$ in the standard edit distance.

\section{Experiments}\label{sec:experiments}
\begin{figure*}[ht]
\begin{center}
\includegraphics[width=0.95\linewidth]{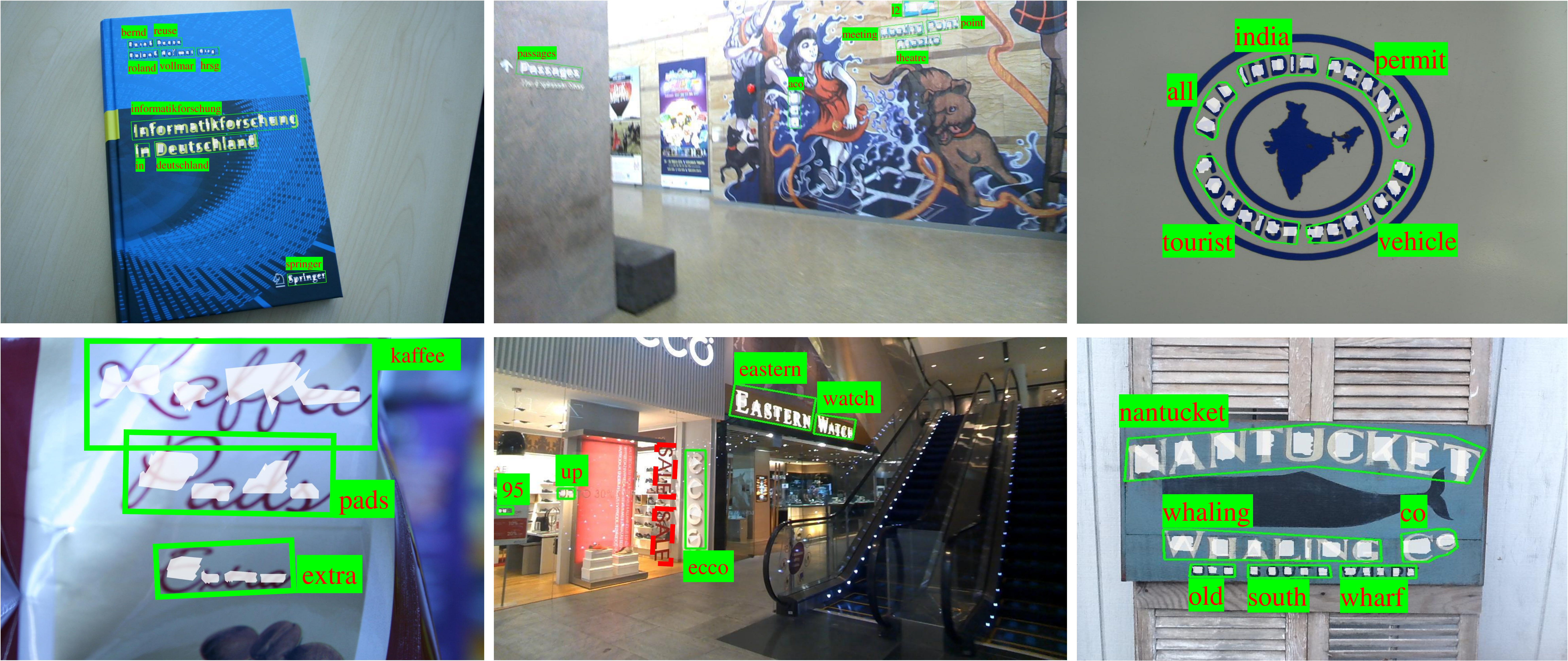}
\end{center}
\caption{Visualization results of ICDAR 2013 (the first column), ICDAR 2015 (the second column) and Total-Text (the last two columns). The dashed red bounding boxes are the false negatives.}
\label{fig:visu}
\end{figure*}

To validate the effectiveness of Mask TextSpotter, we conduct experiments and compare with other state-of-the-art methods on four English datasets and one multi-language dataset for detection/word spotting/end-to-end recognition: a horizontal text set ICDAR2013~\cite{karatzas2013icdar}, two oriented text sets ICDAR2015~\cite{karatzas2015icdar} and COCO-Text~\cite{coco-text/VeitMNMB16}, a curved text set Total-Text~\cite{CK2017}, and a multi-language dataset MLT~\cite{mlt-dataset}. Some results are visualized in Fig.~\ref{fig:visu}. Further, we conduct experiments on scene text recognition benchmarks using the standalone recognition model, to verify the effectiveness of our recognition part.

\subsection{Datasets}
\subsubsection{Synthetic Datasets}
\textbf{SynthText} is a synthetic dataset proposed by \cite{SynthText}, including about $800000$ images. There are a large number of multi-oriented text instances, which are annotated with word-level, character-level rotated bounding boxes, as well as text sequences. All samples in this dataset are used for training the mask text spotter and the standalone recognition model. For the standalone recognition model, the training samples are obtained via cropping text regions from the images.
\\
\noindent\textbf{Synth90k}~\cite{synth90} is a synthetic dataset for scene text recognition. It consists of $9$ million images which cover 90k common English words. Each image is labeled with a ground truth word, without character-level annotations.

\subsubsection{Datasets for Detection and End-to-end Recognition}
\textbf{ICDAR2013} is a dataset proposed in Challenge 2 of the ICDAR 2013 Robust Reading Competition \cite{karatzas2013icdar} which focuses on the horizontal text detection and recognition in natural images. There are $229$ images in the training set and $233$ images in the test set. Besides, the bounding box and the transcription are also provided for each word-level and character-level text instance. 
\\
\textbf{ICDAR2015} is proposed in the ICDAR 2015 Robust Reading Competition \cite{karatzas2015icdar}. Compared to ICDAR2013 which focuses on ``focused text" in particular scenarios, ICDAR2015 is more concerned with the incidental scene text detection and recognition. It contains $1000$ training samples and 500 test images. All training images are annotated with word-level quadrangles as well as corresponding transcriptions. 
\\
\textbf{COCO-Text} consists of $63686$ images. Three versions of the annotations (V1.1， V1.4, and V2.0) are given. It is one of
the challenges of ICDAR 2017 Robust Reading Competition. Though it is evaluated with axis-aligned bounding boxes, the text instances in the images are distributed in various orientations.
\\
\textbf{Total-Text} is a comprehensive scene text dataset proposed by \cite{CK2017}. Except for the horizontal text and oriented text, Total-Text also consists of a lot of curved text. Total-Text contains $1255$ training images and $300$ test images. All images are annotated with polygons and transcriptions at word level. 
\\
\textbf{MLT}~\cite{mlt-dataset} is a multi-language scene text dataset proposed in ICDAR 2017. It consists of 7200 training images, 1800 validation images, and 9000 test images.

\subsubsection{Datasets for Recognition}
\noindent\textbf{IIIT5k-Words} (IIIT5k)~\cite{DBLP:conf/cvpr/MishraAJ12} contains $3000$ test images. There are two lexicons for each image, with a size of 50 and a size of $1k$ respectively. Each lexicon contains the corresponding ground-truth word.

\noindent\textbf{Street View Text} (SVT)~\cite{wang2011end} contains $647$ images of cropped word, which are collected from the Google Street View. A 50-word lexicon is provided for each image like IIIT5k.

\noindent\textbf{ICDAR 2003} (IC03) is cropped and filtered from~\cite{icdar03}. It contains $860$ images, 50-word lexicons for each image, and a lexicon consists of all ground-truth words.

\noindent\textbf{ICDAR 2013} (IC13)~\cite{karatzas2013icdar} is inherited from IC03 with some new images. The test set consists of $1015$ images and no lexicon is given.

\noindent\textbf{ICDAR 2015 Incidental Text} (IC15) are provided by the Task 4.3 of the ICDAR 2015 competition~\cite{karatzas2015icdar}. The images are taken by Google glasses incidentally. There is a large portion of oriented text in this dataset.

\noindent\textbf{SVT-Perspective} (SVTP)~\cite{quy2013recognizing} is similar to SVT. However, most of them are distorted by perspective transformation. The test set contains $639$ cropped images. A 50-word lexicon is provided for each image.

\noindent\textbf{CUTE80} (CUTE)~\cite{risnumawan2014robust} consists of $288$ images. It is a challenging dataset since there are plenty of images with curved text. No lexicon is given.

\begin{table*}[!ht]
\begin{centering}
\caption{The detection results on ICDAR2013 and ICDAR2015. For ICDAR2013, all methods are evaluated under the ``DetEval” evaluation protocol. The short sides of the input image in ``Ours (det only)" and ``Ours'' are set to $1000$.}
\label{tab_detection}
\begin{tabular}{|c|c|c|c|c|c|c|c|c|}
\hline 
\multirow{2}{*}{Method} & \multicolumn{3}{c|}{ICDAR2013} & \multirow{2}{*}{FPS} & \multicolumn{3}{c|}{ICDAR2015} & \multirow{2}{*}{FPS}\tabularnewline
\cline{2-4} \cline{6-8} 
 & precision & recall & f-measure &  & precision & recall & f-measure & \tabularnewline
\hline 
\hline
Zhang \emph{et al.} \cite{zhang2016multi} & 88.0 & 78.0 & 83.0 & 0.5 & 71.0 & 43.0 &  54.0 & 0.5 \tabularnewline
\hline 
CTPN \cite{tian2016detecting} & 93.0 & 83.0  & 88.0 & 7.1 & 74.0 & 52.0 & 61.0 & -  \tabularnewline
\hline 
Seglink \cite{shi2017detecting} MS & 87.7 & 83.0 & 85.3  & 20.6 & 73.1 & 76.8 & 75.0 & - \tabularnewline
\hline 
EAST \cite{zhou2017east} MS & - & - & - & - & 83.3 & 78.3 & 80.7 & -   \tabularnewline
\hline 
SSTD \cite{he2017single} & 89.0 & 86.0  & 88.0 & 7.7 & 80.0 & 73.0 &  77.0 & 7.7 \tabularnewline
\hline 
Wordsup \cite{hu2017wordsup} MS & 93.3 & 87.5 & 90.3 & 2 & 79.3 & 77.0 & 78.2 & 2  \tabularnewline
\hline 
Lyu~\emph{et al.} \cite{lyu2018multi} & 93.3 & 79.4 & 85.8 & 10.4 & \textbf{94.1} & 70.7 & 80.7 & 3.6 \tabularnewline
\hline 
RRD \cite{liao2018rotation} & 88.0 & 75.0 & 81.0 & - & 85.6 & 79.0 & 82.2 & 6.5 \tabularnewline
\hline 
TextSnake \cite{long2018textsnake} & - & - & - & - & 84.9 & 80.4 & 82.6 & 1.1 \tabularnewline
\hline 
Xu~\emph{et al.} \cite{xue2018accurate} & 91.5 & 87.1 & 89.2 & - & -- & -- & -- & -- \tabularnewline
\hline 
He~\emph{et al.} \cite{he2018end} & 91.0 & 88.0 & 90.0 & - & 87.0 & 86.0 & 87.0 & - \tabularnewline
\hline 
FOTS \cite{liu2018fots} & - & - & 88.3 & \textbf{23.9} & 91.0 & 85.17 & \textbf{88.0} & \textbf{7.8} \tabularnewline

\hline 
\hline
Conference version~\cite{LyuLYWB18} &\textbf{95.0}  &88.6  &91.7  &4.6   &91.6  &81.0  &86.0 &4.8 \tabularnewline
\hline
\hline
Ours (det only) &94.1  &88.1  &91.0  &4.6   &85.8   &81.2  &83.4  &4.8 \tabularnewline
\hline 
Ours &94.8  &\textbf{89.5}  &\textbf{92.1}  &3.0 &86.6  &\textbf{87.3}  &87.0 &3.1 \tabularnewline
\hline
\end{tabular}
\par\end{centering}
\end{table*}

\begin{table*}[ht]
\begin{centering}
\caption{Results on ICDAR2013. ``S", ``W" and ``G" mean recognition with strong, weak and generic lexicon respectively. ``MS": testing with multiple scales.}
\label{tab_icdar2013}
\begin{tabular}{|c|c|c|c|c|c|c|c|}
\hline 
\multirow{2}{*}{Method} & \multicolumn{3}{c|}{Word Spotting} & \multicolumn{3}{c|}{End-to-End} & \multirow{2}{*}{FPS}\tabularnewline
\cline{2-7} 
 & S & W & G & S & W & G & \tabularnewline
\hline 
\hline
Jaderberg \emph{et al.} \cite{jaderberg2016reading} & 90.5  & - & 76 & 86.4 & - & - & - \tabularnewline
\hline 
 FCRNall+multi-filt \cite{SynthText} & - & - & 84.7 & - & - & - &  - \tabularnewline
 \hline 
Textboxes \cite{liao2017textboxes} MS  & 93.9  & 92.0  & 85.9 & 91.6 & 89.7 & 83.9 & - \tabularnewline
 \hline 
Deep text spotter \cite{Busta_2017_ICCV} & 92 & 89 & 81 & 89 & 86 & 77 & \textbf{9} \tabularnewline
\hline 
Li \emph{et al.} \cite{Li_2017_ICCV}  &94.2 &92.4 &\textbf{88.2} &91.1 &89.8 &84.6 & 1.1 \tabularnewline
 \hline 
 TextBoxes++~\cite{liao2018textboxes++} MS  &\textbf{96.0} &\textbf{95.0} &87.0 &93.0 &\textbf{92.0} &85.0 &-
 \tabularnewline
 \hline 
 He~\emph{et al.}~\cite{he2018end}  &93.0 &92.0 &87.0 &91.0 &89.0 &86.0 &-
 \tabularnewline
  \hline 
 FOTS~\cite{liu2018fots}  &92.7 &90.7 &83.5 &88.8 &87.1 &80.8 &22.0
 \tabularnewline
\hline 
\hline
Conference version~\cite{LyuLYWB18}  &92.5 &92.0 &\textbf{88.2} &92.2 &91.1 &86.5 &4.8
 \tabularnewline
 \hline 
\hline
 Ours  &92.7 &91.7 &87.7 &\textbf{93.3} &91.3 &\textbf{88.2} &3.1  \tabularnewline
 \hline
\end{tabular}
\par\end{centering}
\end{table*}

\subsection{Implementation Details}
We implement our method in PyTorch\footnote{https://pytorch.org/} and conduct all experiments on a regular workstation with Nvidia Titan Xp GPUs. The model is trained in parallel and evaluated on a single GPU.
\subsubsection{Mask TextSpotter}\label{subsubsec: masktextspotter_implementation}
Different from previous text spotting methods which use two independent models \cite{jaderberg2014deep,liao2017textboxes} (the detector and the recognizer) or alternating training strategy \cite{Li_2017_ICCV}, all subnets of our model can be trained synchronously and end-to-end. The whole training process contains two stages: pre-trained on SynthText and fine-tuned on the real-world data. 

We set the mini-batch to $8$ for all experiments. The batch sizes of RPN and Fast R-CNN are set to $256$ and $512$ per image with a $1:3$ sample ratio of positives to negatives. The batch size of the mask branch is $64$. In the fine-tuning stage, data augmentation and multi-scale training are applied due to the lack of real samples. Specifically, for data augmentation, we randomly rotate the input pictures in a certain angle range of $[-30^\circ, 30^\circ]$. Some other augmentation tricks, such as modifying the hue, brightness, contrast randomly, are also used following \cite{liu2016ssd}. For multi-scale training, the shorter sides of the input images are randomly resized to five scales $(600, 800, 1000, 1200, 1400)$ randomly. Besides, following \cite{Li_2017_ICCV}, extra $1162$ images (SCUT) from \cite{zhong2016deeptext} are also used as training samples. 
For each mini-batch, we sample the images from the training datasets with a fixed sample ratio, which is set to $2:2:2:1:1$ for SynthText, ICDAR2013, ICDAR2015, Total-Text, and SCUT respectively.

We optimize our model using SGD with a weight decay of $0.001$ and momentum of $0.9$. In the pre-training stage, we train our model for $270k$ iterations, with an initial learning rate of $0.01$. Then the learning rate is decayed to a tenth at the $100k$ and $200k$ iteration. In the fine-tuning stage, the initial learning rate is set to $0.001$, and then be decreased to a tenth at the $100k$ iteration. The fine-tuning process is terminated at the $150k$ iteration.

In the inference stage, we evaluate all English datasets with a single model, while the scales of the input images depend on different datasets. After NMS, $1000$ proposals are fed into Fast R-CNN. False alarms and redundant candidate boxes are filtered out by Fast R-CNN and NMS respectively. The kept candidate boxes are input to the mask branch to generate the text instance maps, the character maps, and the text sequence. Finally, the text instance bounding boxes and sequences are generated from the predicted maps.

\subsubsection{Standalone Recognition Model}
The standalone recognition model is trained on synthetic data only, with the mini-batch of $128$. It is optimized using ADAM with a weight decay of $1e^{-4}$ and momentum of $0.9$. The base learning rate is set to $2e^{-4}$ and then decayed to a tenth after $2$ epochs. The training is terminated after $3$ epochs. Data augmentation and multi-scale training technology are also applied as mentioned in Sec.~\ref{subsubsec: masktextspotter_implementation}. In detail, the input images are resized to $32 \times 128$, $48 \times 192$, $64 \times 256$ randomly. Different from Mask TextSpotter, we add an extra label which represents non-alphanumeric characters in the standalone recognition model.

In the inference period, the input images are  resized to $h \times w$, where $h$ and $w$ are the height and the width respectively. By default, $h$ is set to $64$ and $w$ is calculated by keeping the aspect ratio of the original image. To avoid too small width of the input image, we set the minimum width as $256$, which means $w = \max(w, 256)$.

\begin{table*}[ht]
\begin{centering}
\caption{Results on ICDAR2015. ``S", ``W" and ``G" mean recognition with strong, weak and generic lexicon respectively. ``MS": testing with multiple scales.}
\label{tab_icdar2015}
\begin{tabular}{|c|c|c|c|c|c|c|c|}
\hline 
\multirow{2}{*}{Method} & \multicolumn{3}{c|}{Word Spotting} & \multicolumn{3}{c|}{End-to-End} & \multirow{2}{*}{FPS}\tabularnewline
\cline{2-7} 
 & S & W & G & S & W & G & \tabularnewline
\hline 
\hline
Baseline OpenCV3.0 + Tesseract\cite{karatzas2015icdar}  & 14.7 & 12.6 & 8.4 & 13.8  & 12.0  & 8.0 & - \tabularnewline
\hline 
TextSpotter \cite{neumann2016real} & 37.0  & 21.0 & 16.0 & 35.0 & 20.0 & 16.0 & 1 \tabularnewline
\hline 
Stradvision \cite{karatzas2015icdar} & 45.9  & - & - & 43.7 & - & - & - \tabularnewline
\hline 
TextProposals + DictNet \cite{gomez2017textproposals,synth90} & 56.0 & 52.3 & 49.7 & 53.3 & 49.6 & 47.2 &  0.2 \tabularnewline
\hline 
HUST\_MCLAB \cite{shi2017detecting,shi2017end} & 70.6 & - & - & 67.9 & - & - &  - \tabularnewline
 \hline 
Deep text spotter \cite{Busta_2017_ICCV} & 58.0 & 53.0 & 51.0 & 54.0 & 51.0 & 47.0 & \textbf{9.0} \tabularnewline
 \hline 
 TextBoxes++~\cite{liao2018textboxes++} MS  &76.5 &69.0 &54.4 &73.3 &65.9 &51.9 &-
 \tabularnewline
 \hline 
 He~\emph{et al.}~\cite{he2018end}  &\textbf{85.0} &\textbf{80.0} &65.0 &82.0 &77.0 &63.0 &-
 \tabularnewline
  \hline 
 FOTS~\cite{liu2018fots}  &84.7 &79.3 &63.3 &81.1 &75.9 &60.8 & 7.5
 \tabularnewline
\hline 
\hline
 Conference version~\cite{LyuLYWB18} (720)  &71.6  &63.9  &51.6  &71.3  &62.5  &50.0  &6.9   \tabularnewline
\hline
 Conference version~\cite{LyuLYWB18} (1000)  &77.7  &71.3  &58.6  &77.3  &69.9  &60.3  &4.8   \tabularnewline
\hline
 Conference version~\cite{LyuLYWB18} (1600)  &79.3  &74.5  &64.2  &79.3  &73.0  &62.4  &2.6   \tabularnewline
\hline
\hline
 Ours (720)  &74.1  &69.7  &64.1  &74.2  &69.2  &63.5  &3.8   \tabularnewline
\hline
 Ours (1000)  &81.4  &76.8  &71.5  &82.0  &76.6  &71.1  &3.1   \tabularnewline
\hline
 Ours (1600)  &82.4  &78.1  &\textbf{73.6}  &\textbf{83.0}  &\textbf{77.7}  &\textbf{73.5}  &2.0   \tabularnewline
\hline
\end{tabular}
\par\end{centering}
\end{table*}

\begin{table*}[ht]
\begin{centering}
\caption{Detection and end-to-end results on COCO-Text.``AP" is short for average precision. Methods with ``*'' are evaluated using V1.1 annotations. ``MS" means testing with multiple scales. Methods with ``**" are results on the competition website. ( http://rrc.cvc.uab.es/?ch=5)}
\label{table:cocotext}
\begin{tabular}{|c|c|c|c|c|c|c|c|}
\hline
\multirow{2}{*}{Method} & \multicolumn{3}{c|}{Detection}                & \multicolumn{4}{c|}{End-to-End}                \\ \cline{2-8} 
                        & precision     & recall        & f-measure     & precision     & recall        & f-measure  & AP     \\ \hline
Baseline A*~\cite{coco-text/VeitMNMB16}             & 83.8          & 23.3          & 36.5          & \textbf{68.4} & 28.3          & 40     & -        \\ \hline
Baseline B*~\cite{coco-text/VeitMNMB16}              & 59.7          & 10.7          & 19.1          & 9.97          & \textbf{54.5} & 16.9     & -      \\ \hline
Baseline C*~\cite{coco-text/VeitMNMB16}             & 18.6          & 4.7           & 7.5           & 1.7           & 4.2           & 2.4      & -      \\ \hline
EAST*~\cite{zhou2017east}                   & 50.4          & 32.4          & 39.5          & -             & -             & -        & -      \\ \hline
WordSup*~\cite{hu2017wordsup}                & 45.2          & 30.9          & 36.8          & -             & -             & -       & -       \\ \hline
SSTD*~\cite{he2017single}                   & 46            & 31            & 37            & -             & -             & -        & -      \\ \hline
UM**                     & 47.6          & \textbf{65.5} & 55.1          & -             & -             & -      & -        \\ \hline
TDN\_SJTU\_v2**             & 62.4          & 54.3          & 51.8          & -             & -             & -    & -          \\ \hline
Text\_Detection\_DL**       & 60.1          & 61.8          & 61.4          & -             & -             & -     & -         \\ \hline
WPS**       & -          & -         & -          & -             & -             & -     & 18.8         \\ \hline
Foo \& Bar**       & -          & -         & -          & -             & -             & -     & 27.0         \\ \hline
Tencent-DPPR Team \& USTB-PRIR**       & -          & -         & -          & -             & -             & -     & 43.6         \\ \hline
RRD~\cite{liao2018rotation} MS   & 64.0  & 57.0          & 61.0          & -             & -             & -      & -        \\ \hline
Lyu \emph{et al.}~\cite{lyu2018multi}            & \textbf{72.5}  & 52.9          & 61.1          & -             & -             & -       & -       \\ \hline
Ours                    & 66.8          & 58.3          & \textbf{62.3} & 65.8          & 37.3          & \textbf{47.61}   & 23.9   \\ \hline
\end{tabular}
\par\end{centering}
\end{table*}

\subsection{Horizontal Text}
We evaluate our model on ICDAR2013 dataset to verify its effectiveness in detecting and recognizing horizontal text. We resize the shorter sides of all input images to $1000$ and evaluate the results online.

The results of our model are listed and compared with other state-of-the-art methods in Table~\ref{tab_detection} and Table~\ref{tab_icdar2013}. 
For the detection task, our method achieves state-of-the-art results. Concretely, for detection, though evaluated at a single scale, our method outperforms some previous methods which are evaluated at multi-scale setting, such as \cite{hu2017wordsup}. For the word spotting and the end-to-end recognition tasks, our method is comparable to the previous best methods even if some of them are tested with multiple scales. Our method performs better when the given lexicon is the generic lexicon (containing 90k words), whose size is much larger than the strong (containing 50 words) or weak lexicon (containing hundreds of words). This means that our method is less reliable to lexicons.

In Table~\ref{tab_icdar2013}, the current results are slightly lower than the conference version in some tasks (2 of 6 tasks, the gaps are 0.3 and 0.5 percent respectively). Since ICDAR2013 contains only 233 testing images, whose size is relatively small, it is a natural disturbance for such a small gap. We can see that the improvements for the other 4 tasks (the gaps are 0.2, 1.1, 0.2, 1.7 percent respectively) are more significant than the decreasing.

\begin{figure*}[ht]
\begin{center}
\includegraphics[width=0.95\linewidth]{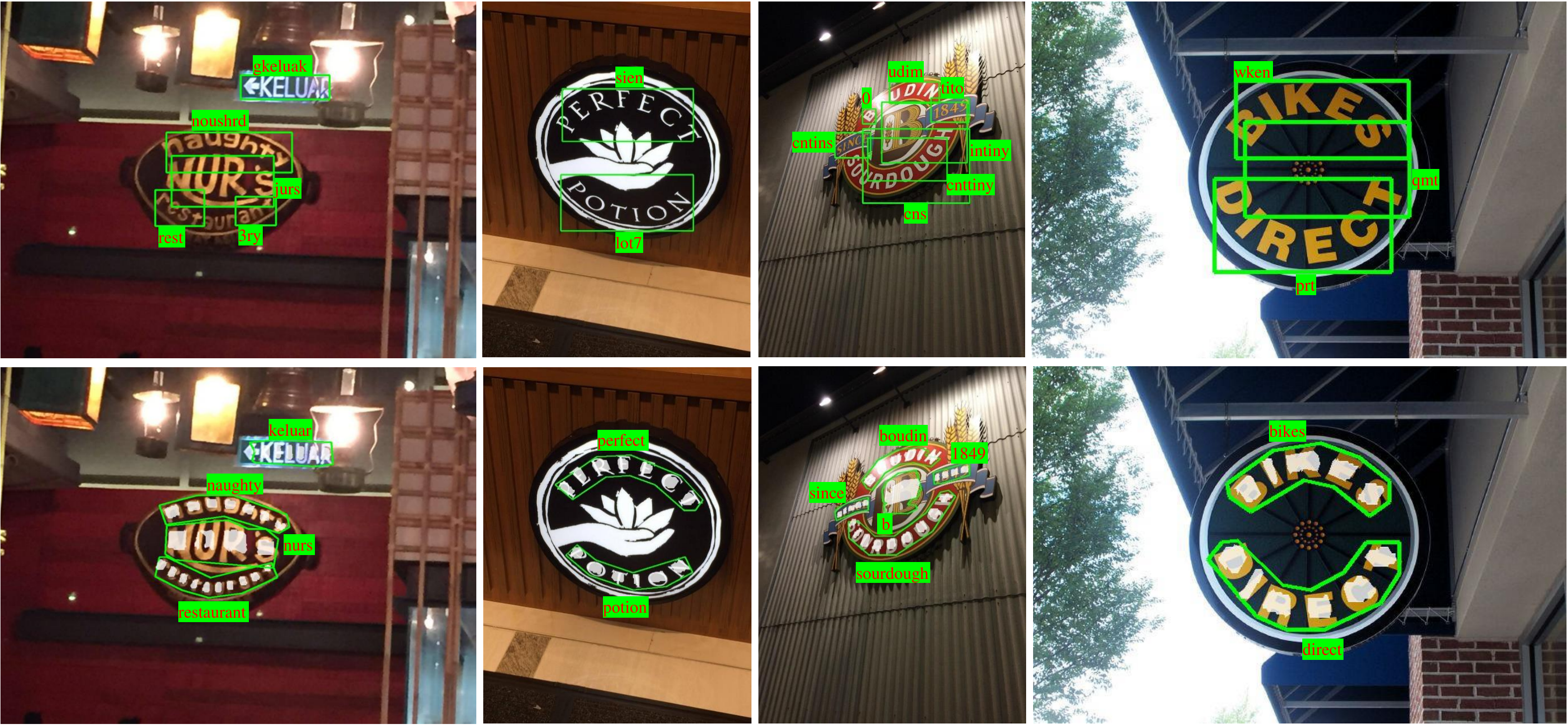}
\end{center}
\caption{Qualitative comparisons on Total-Text without lexicon. Top: results of TextBoxes~\cite{liao2017textboxes}; Bottom: results of ours.}
\label{fig:visu_compare}
\end{figure*}

\subsection{Oriented Text}
We verify the superiority of our method on oriented text by conducting experiments on ICDAR2015. We input the images with three different scales: the original scale ($720 \times 1280$) and two larger scales where shorter sides of the input images are $1000$ and $1600$ due to a lot of small text instance in ICDAR2015. We evaluate our method online and compare it with other methods in Table~\ref{tab_detection} and Table~\ref{tab_icdar2015}. 

For the detection task, our method achieves the f-measure of $87\%$, which is comparable to the previous state-of-the-art method~\cite{liu2018fots}. Moreover, our method achieves the best recall among all methods in Table~\ref{tab_detection}.
For the word spotting and the end-to-end tasks, our method outperforms the previous state of the arts~\cite{he2018end} by $\mathbf{8.6}$ percents and $\mathbf{10.5}$ percents when the provided lexicon is the generic lexicon. When the given lexicon is the strong or weak lexicon, our method also achieves comparable results. The experimental results prove again that our method relies less on lexicons.

We also evaluate our method on COCO-Text~\cite{coco-text/VeitMNMB16,gomez2017icdar2017} to verify its universality. The detection task in Table~\ref{table:cocotext} is evaluated on the ICDAR 2017 Robust Reading Challenge on COCO-Text~\cite{gomez2017icdar2017} with the annotations V1.4 for a fair comparison with previous methods. The end-to-end task is evaluated with the recent annotations V2.0\footnote{https://bgshih.github.io/cocotext}. Following~\cite{lyu2018multi}, we also do not train our model with the training set of COCO-Text. As shown in Table~\ref{table:cocotext}, our method achieves comparable or state-of-the-art performance on both the detection and the end-to-end recognition tasks. 
Note that the methods with ``**" are in the competition website, which may use a lot of tricks such as extra training data, multi-scale testing, large backbone, language model, model ensemble, \textit{et al}. Thus, it is unfair to directly compare with them. We report the average precision on the end-to-end task for reference only.

\begin{table*}[ht]
\begin{centering}
\caption{Results on Total-Text. ``None" means recognition without any lexicon. ``Full" lexicon contains all words in test set. Note that the official evaluation code is updated with some bugs fixed, thus some of the results are different from the conference version.
}
\label{tab_total}
\begin{tabular}{|c|c|c|c|c|c|c|c|c|}
\hline
\multirow{3}{*}{Method} & \multicolumn{6}{c|}{Detection}                                                              & \multicolumn{2}{c|}{End-to-end}               \\ \cline{2-9} 
                        & \multicolumn{3}{c|}{Det\_eval}                & \multicolumn{3}{c|}{PASCAL}                 & \multirow{2}{*}{None} & \multirow{2}{*}{Full} \\ \cline{2-7}
                        & precision     & recall        & f-measure     & precision   & recall        & f-measure     &                       &                       \\ \hline
TextBoxes~\cite{liao2017textboxes}              & 47.2           & 42.5           & 44.7           & 52.8        & 49.7          & 51.2          & 36.3                  & 48.9                  \\ \hline
FTSN~\cite{Dai2017}                    & -             & -             & -             & 84.7        & 78            & 81.3          & -                     & -                     \\ \hline
Conference version~\cite{LyuLYWB18}       & 79.1 & \textbf{77.9}          & \textbf{78.5}           & 87.7 & 80.5         & 83.9         & 52.9                  & 71.8                  \\ \hline
Ours                    & \textbf{81.8}          & 75.4 & \textbf{78.5} & \textbf{88.3}        & \textbf{82.4} & \textbf{85.2} & \textbf{65.3}         & \textbf{77.4}         \\ \hline
\end{tabular}
\par\end{centering}
\end{table*}

\subsection{Curved Text}
Detecting and recognizing arbitrary text (e.g. curved text) is a huge superiority of our method beyond other methods. We conduct experiments on Total-Text to verify the robustness of our method in detecting and recognizing curved text. Similarly, we input the test images with the short edges resized to $1000$. Since the official evaluation protocol is updated, we use the updated Python scripts\footnote{https://github.com/cs-chan/Total-Text-Dataset/tree/master/
Evaluation Protocol} provided by the official to evaluate the detection task. In Table~\ref{tab_total}, ``PASCAL" uses a polygon IoU threshold of $0.5$. ``Det\_eval" adopts two stricter thresholds, where ``tr" and ``tp" are set to $0.7$ and $0.6$ respectively.
Note that although \cite{long2018textsnake} reported detection results on Total-Text, it used a different evaluation protocol. Thus, for a fair comparison, we do not list their performance in Table~\ref{tab_total}, and just report the results here for reference (precision: $82.7\%$, recall: $74.5\%$ f-measure: $78.4\%$). 
The evaluation protocol of end-to-end recognition follows ICDAR 2015 while changing the representation of polygons from four vertexes to an arbitrary number of vertexes in order to handle the polygons of arbitrary shapes.

To compare with other methods, we also trained a model in TextBoxes~\cite{liao2017textboxes} using the official code\footnote{https://github.com/MhLiao/TextBoxes} with the same training data. As shown in Fig.~\ref{fig:visu_compare}, our method possesses an apparent advantage over TextBoxes on both detecting and recognizing curved text. Moreover, the results in Table~\ref{tab_total} show that our method exceeds \cite{Dai2017} by $\mathbf{3.9}$ percents in detection and outperforms \cite{liao2017textboxes} by at least $\mathbf{28.5\%}$ in end-to-end recognition. Benefiting from the integration of both detection and recognition, our method achieves better performance on the detection task.
As for recognition, our method is more suitable to recognize text sequences distributed in two-dimensional space (such as curves) and outperforms the sequence recognition network used in \cite{liao2017textboxes} which is designed for one-dimensional sequences. 

Compared with our conference version~\cite{LyuLYWB18}, our method achieves $\mathbf{12.4}$ percents (without lexicon) and $\mathbf{5.6}$ percents (with lexicon) performance gain in the end-to-end recognition task. Slight improvements are also achieved in the detection task. 
The performance gains demonstrate that our newly proposed SAM can significantly improve the recognition accuracy and marginally enhance the detection performance. One reason is that SAM, which reads the text in a global view, is complementary to the character segmentation module which predicts the characters locally. Another reason is that SAM does not require character annotations so it can use more real-world word-level annotations to supervise. Moreover, the improvements on the recognition slightly benefit the detection.

\subsection{Multi-Language End-to-End Recognition}
We conduct experiments on MLT dataset to prove that our method is robust when the number of character classes is large (more than 7000 character classes). We follow Busta et al.~\cite{e2e-mlt} with the same backbone (ResNet-34), same training data provided by Busta et al.~\cite{e2e-mlt}, for fair comparison. Since there are no character-level annotations in the training data, we disable the character segmentation branch in this experiment. As shown in Table.~\ref{table:mlt}, our method outperforms Busta et al.~\cite{e2e-mlt} by a large margin, which demonstrates that it can deal with a large number of character classes.

\begin{table}[ht]
\caption{Results on MLT dataset. ``Det-R": detection recall; ``E2E-R": end-to-end recognition recall; ``E2E-R ED1": end-to-end recognition recall where the edit distance is small or equal to 1; ``Det-R": detection recall; ``P": precision. ``2+" and ``3+" mean that words whose length are large than 2 and 3 are counted respectively. ``*" means ignoring the difficult labels in the evaluation.
}
\label{table:mlt}
\centering
\begin{tabular}{|c|c|c|c|c|}
\hline
\multirow{2}{*}{Method} & \multicolumn{4}{c|}{MLT Validation Set}                    \\ \cline{2-5} 
                        & Det-R & E2E-R & E2E-R ED1 & P \\ \hline
Busta et al.~\cite{e2e-mlt} 2+         & 68.4             & 42.9       & 55.5           & 53.7      \\ \hline
Ours 2+                 & \textbf{80.0}             & \textbf{47.9}       & \textbf{71.3}           & \textbf{68.3}      \\ \hline
\hline
Busta et al.~\cite{e2e-mlt} 3+         & 69.5             & 43.3       & 59.9           & 59.7      \\ \hline
Ours 3+                 & \textbf{82.8}             & \textbf{48.5}       & \textbf{74.2}           & \textbf{60.5}      \\ \hline
\hline
Ours 2+*                & 80.0             & 47.9       & 71.3           & 75.2     \\ \hline
Ours 3+*                & 82.8             & 48.5       & 74.2           & 72.8      \\ \hline
\end{tabular}
\end{table}

\subsection{Speed}
The speed comparisons are shown in Table~\ref{tab_icdar2015}. We can see that though our model is not the fastest, the speed of our method is comparable to previous methods. Specifically, it can run at $3.8$ FPS, $3.1$ FPS, and $2.0$ FPS with the input scale of $720 \times 1280$, $1000 \times 1778$, and $1600 \times 2844$ respectively.
For the input scale of  $720 \times 1280$, it takes about $0.20$ second for the detection and $0.06$ second for the recognition averagely.

\subsection{Ablation Experiments}
\begin{table*}[ht]
\begin{centering}
\caption{Ablation experimental results. ``(a)" means without character-level annotations from the real images; ``(b)" means without weighted edit distance. $\Delta$ means the variation compared to the original version.}
\label{tab_discussion}
\begin{tabular}{|c|c|c|c|c|c|c|c|c|c|c|c|c|}
\hline 
\multirow{3}{*}{Settings} & \multicolumn{6}{c|}{ICDAR2013} & \multicolumn{6}{c|}{ICDAR2015}\tabularnewline
\cline{2-13} 
 & \multicolumn{3}{c|}{Word Spotting} & \multicolumn{3}{c|}{End-to-End} & \multicolumn{3}{c|}{Word Spotting} & \multicolumn{3}{c|}{End-to-End}\tabularnewline
\cline{2-13} 
 & S & W & G & S & W & G & S & W & G & S & W & G\tabularnewline
\hline 
\hline
Conference version &\textbf{92.5}  &\textbf{92.0}  &\textbf{88.2} &\textbf{92.2}  &\textbf{91.1}  &\textbf{86.5}  &\textbf{79.3}  &\textbf{74.5}  &\textbf{64.2}  &\textbf{79.3}  &\textbf{73.0}  &\textbf{62.4} \tabularnewline
Conference version (a) &91.8  &90.3  &85.9  &90.7  &89.4  &84.6  &76.9  &71.6  &61.6  &76.6  &69.9 &59.8 \tabularnewline
Conference version (b) &91.4  &90.5  &84.3  &91.3  &89.9  &83.8  &75.9  &67.5  &56.8  &76.1  &67.1  &56.7 \tabularnewline
\hline 
\hline
Ours &\textbf{92.7}  &\textbf{91.7}  &\textbf{87.7} &\textbf{93.3}  &\textbf{91.3}  &\textbf{88.2}  &82.4  &78.1  &\textbf{73.6}  &83.0  &77.7 &\textbf{73.5} \tabularnewline
Ours (a) &92.0 &91.0 &87.6 &92.6 &90.4 &87.4 &\textbf{82.7} &\textbf{78.3} &72.5 &\textbf{83.3} &\textbf{77.9} &72.3 \tabularnewline
Ours (b) &92.3  &91.0  &87.7  &93.0  &90.5  &88.0  &81.9  &77.7  &72.2  &82.1  &77.0  &72.0 \tabularnewline
\hline 
\hline
Conference version (a)$\Delta$ &\textbf{-0.7} &-1.7 &-2.3 &-1.5 &-1.7 &-1.9 &-2.4 &-2.9 &-2.6 &-2.7 &-3.1 &-2.6 \tabularnewline
Ours (a)$\Delta$ &\textbf{-0.7} &\textbf{-0.7} &\textbf{-0.1} &\textbf{-0.7} &\textbf{-0.9} &\textbf{-0.8} &\textbf{+0.3} &\textbf{+0.2} &\textbf{-1.1} &\textbf{+0.3} &\textbf{+0.2} &\textbf{-1.2} \tabularnewline
\hline 
\hline
Conference version (b)$\Delta$ &-1.1  &-1.5  &-3.9  &-0.9  &-1.2  &-2.7  &-3.4  &-7.0  &-7.4  &-3.2  &-5.9  &-5.7 \tabularnewline
Ours (b)$\Delta$ &\textbf{-0.4}  &\textbf{-0.7}  &\textbf{0}  &\textbf{-0.3}  &\textbf{-0.8}  &\textbf{-0.2}  &\textbf{-0.5}  &\textbf{-0.4}  &\textbf{-1.4}  &\textbf{-0.9}  &\textbf{-0.7}  &\textbf{-1.5 }\tabularnewline
\hline 
\end{tabular}
\par\end{centering}
\end{table*}

\begin{table}[ht]
\caption{Experiments on Backbone and RoI size of the mask branch. The short sides of the input images are set to $1600$.}
\centering
\begin{tabular}{|c|c|c|c|c|c|}
\hline
\multirow{2}{*}{Backbone} & \multirow{2}{*}{RoI size} & \multicolumn{3}{c|}{ICDAR2015 End-to-End}   &  \multirow{2}{*}{FPS}  \\ \cline{3-5} 
                          &                           & S             & W             & G     &   \\ \hline
ResNet-34                      & 16 $\times$ 64                     & 83.0 & 77.6          & 72.6    &  \textbf{2.3}    \\ \hline
ResNet-50                      & 8 $\times$ 32                      & 82.1          & 76.7          & 71.3    &  2.1    \\ \hline
ResNet-50                      & 16 $\times$ 128                    & 82.7          & 77.0          & 71.6     &  1.7   \\ \hline
ResNet-50                      & 32 $\times$ 32                    & 82.7          & 76.9          & 73.2     &  2.0   \\ \hline
ResNet-50                      & 16 $\times$ 64                     & \textbf{83.0}          & \textbf{77.7} & \textbf{73.5} & 2.0 \\ \hline
\end{tabular}
\label{tab:roi}
\end{table}

\noindent\textbf{With or Without the Recognition Part} We train a model named ``Ours (det only)" which removes the recognition part from the original network to explore the advantage of training detection and recognition jointly. As shown in Table~\ref{tab_detection}, the detection results of ``Ours" exceed ``Ours (det only)" by $1.1\%$ and $3.6\%$ on ICDAR2013 and ICDAR2015 respectively, which demonstrate that the detection task can benefit from the recognition task when jointly training.
\\
\textbf{With or Without Real-world Character Annotations} The experiments without real-world character-level annotations are also conducted. As shown in Table~\ref{tab_discussion}, although ``Ours (a)" is trained without any real-world character-level annotation, it still achieves competitive performances. More specifically, for horizontal text (ICDAR2013), it decreases ``Ours", which is trained with a few real-world character-level annotations, by $0.1\%-0.9\%$ on various settings; on ICDAR2015, ``Ours (a)" even achieves better results than ``Ours" on some settings, which demonstrates that our method does not highly rely on the real-world character-level annotations.
Compared with the corresponding conference version which decreases $0.7\%-2.3\%$ and $2.4\%-3.1\%$ on ICDAR2013 and ICDAR2015, our method achieves almost equal performance without real-world character-level annotations.
\\
\textbf{With or Without Weighted Edit Distance} We conduct experiments to verify the effectiveness of our proposed weighted edit distance. The methods of using original edit distance are named with ``(b)". As shown in Table~\ref{tab_discussion}, the weighted edit distance can boost the performance by at most $7.4$ percents among all tasks in the conference version. However, it can only improve the performance by at most $1.5$ percents in our method, as our method without weighted edit distance (``Ours (b)") has already surpassed the conference version with weighted edit distance (``Conference version").
Nevertheless, the weighted edit distance is effective since it achieves better results than the original edit distance, on both the conference version and our method.
\\
\textbf{Backbone \& RoI Size} We conduct ablation study on the backbone and the RoI sizes of the mask branch. As shown in Table~\ref{tab:roi}, the ResNet-34 backbone achieves comparable but a little lower performance than the ResNet-50 Backbone, which indicates that our model can gain better speed by applying a smaller backbone. As for the RoI size of the mask branch, ``$16 \times 64$" achieves the best performance.

\begin{table*}[ht]
\caption{Scene text recognition results. ``50'', ``1k'', ``Full'' are lexicons. ``0'' means no lexicon. ``90k'' and ``ST'' are the Synth90k and the SynthText datasets, respectively. ``Private'' means private training data. $^*$The results of ASTER on SVT were clarified and updated in the web page\protect\footnotemark.}
\renewcommand{\arraystretch}{1.0}
\centering
\resizebox{\textwidth}{!}{%
\begin{tabular}{|l|l|ccc|cc|ccc|c|c|c|c|}
\hline 
\multirow{2}{*}{Methods} & \multirow{2}{*}{Backbone, Data} & \multicolumn{3}{c|}{IIIT5k} & \multicolumn{2}{c|}{SVT} & \multicolumn{3}{c|}{IC03} & IC13 & IC15 & SVTP & CUTE\tabularnewline
\cline{3-14} 
 &  & 50 & 1k & 0 & 50 & 0 & 50 & Full & 0 & 0 & 0 & 0 & 0\tabularnewline
\hline 
Wang \emph{et al.} \cite{wang2011end} & - & - & - & - & 57.0 & - & 76.0 & 62.0 & - & - & - & - & -\tabularnewline
Mishra \emph{et al.} \cite{DBLP:conf/bmvc/MishraAJ12} & - & 64.1 & 57.5 & - & 73.2 & - & 81.8 & 67.8 & - & - & - & - & -\tabularnewline
Wang \emph{et al.} \cite{WangWCN12} & - & - & - & - & 70.0 & - & 90.0 & 84.0 & - & - & - & - & -\tabularnewline
Bissacco \emph{et al.} \cite{bissacco2013photoocr} & - & - & - & - & - & - & 90.4 & 78.0 & - & 87.6 & - & - & -\tabularnewline
Almazan \emph{et al.} \cite{AlmazanGFV14} & - & 91.2 & 82.1 & - & 89.2 & - & - & - & - & - & - & - & -\tabularnewline
Yao \emph{et al.} \cite{yao2014strokelets} & - & 80.2 & 69.3 & - & 75.9 & - & 88.5 & 80.3 & - & - & - & - & -\tabularnewline
Rodr{\'{\i}}guez{-}Serrano \emph{et al.} \cite{Rodriguez-Serrano15} & - & 76.1 & 57.4 & - & 70.0 & - & - & - & - & - & - & - & -\tabularnewline
Jaderberg \emph{et al.} \cite{jaderberg2014deep} & - & - & - & - & 86.1 & - & 96.2 & 91.5 & - & - & - & - & -\tabularnewline
Su and Lu \cite{SuL14} & - & - & - & - & 83.0 & - & 92.0 & 82.0 & - & - & - & - & -\tabularnewline
Gordo \cite{Gordo14} & - & 93.3 & 86.6 & - & 91.8 & - & - & - & - & - & - & - & -\tabularnewline
Jaderberg \emph{et al.} \cite{jaderberg2016reading} & VGG, 90k & 97.1 & 92.7 & - & 95.4 & 80.7 & 98.7 & \textbf{98.6} & 93.1 & 90.8 & - & - & -\tabularnewline
Jaderberg \emph{et al.} \cite{JaderbergSVZ14b} & VGG, 90k & 95.5 & 89.6 & - & 93.2 & 71.7 & 97.8 & 97.0 & 89.6 & 81.8 & - & - & -\tabularnewline
Shi \emph{et al.} \cite{shi2017end} & VGG, 90k & 97.8 & 95.0 & 81.2 & 97.5 & 82.7 & 98.7 & 98.0 & 91.9 & 89.6 & - & - & -\tabularnewline
Lee \emph{et al.} \cite{lee2016recursive} & VGG, 90k & 96.8 & 94.4 & 78.4 & 96.3 & 80.7 & 97.9 & 97.0 & 88.7 & 90.0 & - & - & -\tabularnewline
Yang \emph{et al.} \cite{yang2017learning} & VGG, Private & 97.8 & 96.1 & - & 95.2 & - & 97.7 & - & - & - & - & 75.8 & 69.3\tabularnewline
Cheng \emph{et al.} \cite{ChengBXZPZ17} & ResNet, 90k+ST & 99.3 & 97.5 & 87.4 & 97.1 & 85.9 & 99.2 & 97.3 & 94.2 & 93.3 & 70.6 & - & -\tabularnewline
Cheng \emph{et al.} \cite{cheng2018aon} & self-design, 90k+ST & 99.6 & 98.1 & 87.0 & 96.0 & 82.8 & 98.5 & 97.1 & 91.5 & - & 68.2 & 73.0 & 76.8\tabularnewline
Bai \emph{et al.} \cite{bai2018edit} & ResNet, 90k+ST & 99.5 & 97.9 & 88.3 & 96.6 & 87.5 & 98.7 & 97.9 & 94.6 & 94.4 & 73.9 & - & -\tabularnewline
$^*$ASTER-A~\cite{shi2018aster} & ResNet, 90k & 98.7 & 96.3 & 83.2 & 96.1 & 81.6 & 99.1 & 97.6 & 92.4 & 89.7 & 68.9 & 75.4 & 67.4\tabularnewline
$^*$ASTER-B~\cite{shi2018aster} & ResNet, 90k+ST & 99.6 & 98.8 & 93.4 & 97.4 & 89.5 & 98.8 & 98.0 & 94.5 & 91.8 & 76.1 & 78.5 & 79.5\tabularnewline
\hline 
Ours-segmentation & ResNet, ST &99.7  &99.1  &94.0  & 98.0 &87.2  &\textbf{99.3}  &98.0  &93.1  &92.3  &73.8  &76.3  &82.6 \tabularnewline
Ours-SAM-without-PE & ResNet, ST &99.2  &97.8  &90.1  &97.1  &86.1  &97.9  &95.5  &86.0  &88.4  &72.5  &75.5  &78.1 \tabularnewline
Ours-SAM-A & ResNet, ST &99.3  &97.8  &91.1  &97.7  &87.0  &98.6  &97.1  &90.9  &90.8  &73.0  &76.4  &84.0 \tabularnewline
Ours-SAM-B & ResNet, 90k+ST &99.4  &98.6  &93.9  &98.6  &90.6  &98.8  &98.0  &\textbf{95.2}  &95.3  &77.3  &82.2  &87.8 \tabularnewline
Ours-seg-SAM & ResNet, 90k+ST & \textbf{99.8} & \textbf{99.3} & \textbf{95.3} & \textbf{99.1} & \textbf{91.8} & 99.0 & 97.9 & 95.0 & \textbf{95.3} & \textbf{78.2} & \textbf{83.6} & \textbf{88.5}\tabularnewline
\hline 
\end{tabular}
}
\label{tab:recognition}
\end{table*}

\subsection{Experiments on the Standalone Recognition Model}
\begin{figure}[ht]
\begin{center}
\includegraphics[width=1.0\linewidth]{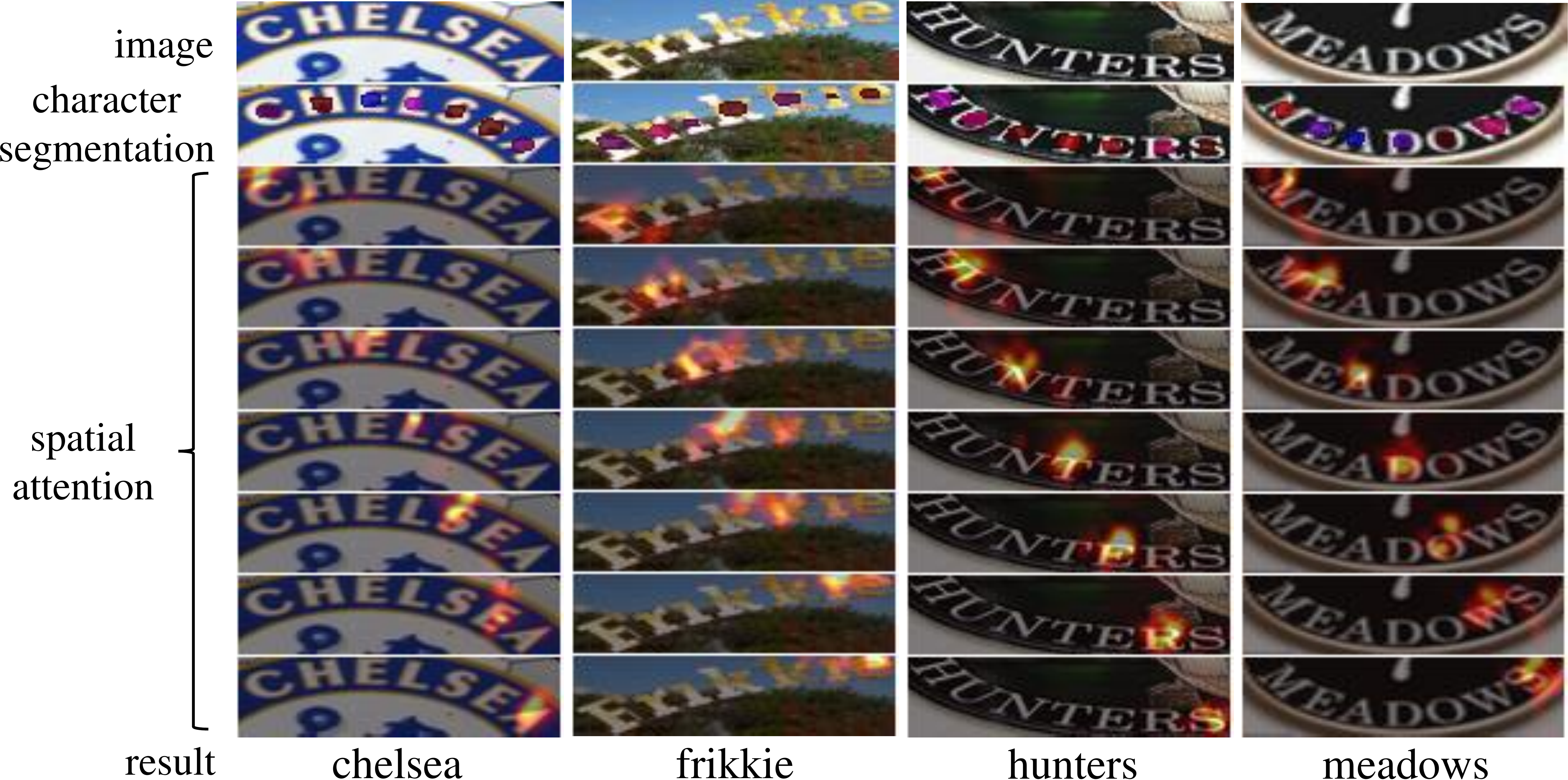}
\end{center}
\caption{Visualization results of the character segmentation maps and the spatial attention weights. Best viewed in color.}
\label{fig:attention_visu}
\end{figure}

We build a standalone recognition model to verify the superiority of our recognition part of Mask TextSpotter. Some visualization results of the character segmentation maps and the spatial attention weights are shown in Fig.~\ref{fig:attention_visu}. From the visualization of the spatial attention, we can see that the model focuses on the corresponding areas at each predicting step. Note that for a fair comparison with previous scene text recognition methods, weighted edit distance is not applied in the standalone recognition model.

\subsubsection{Comparison with SOTA}
 As shown in Table~\ref{tab:recognition}, our model outperforms the previous state-of-the-art method ASTER~\cite{shi2018aster} on all $12$ tasks among $7$ scene text recognition benchmarks. Concretely, our model surpasses ASTER by $\mathbf{5.1\%}$ and $\mathbf{9.0\%}$ on SVTP and CUTE respectively, which demonstrates that it is superior on hugely distorted shapes, such as perspective shape and curved shape. The experimental results strongly prove the effectiveness and robustness of our recognition model.

\subsubsection{Comparison with Related Recognition Methods}
Compared with the rectification-based recognition method~\cite{shi2018aster} which rectifies the shape of the text before recognition, our model directly reads the text in two-dimensional space. Thus, the recognition results of ours do not limit to the effectiveness of the rectification. 
For fair comparisons, we compare ``ASTER-B" with ``Ours-SAM-B", which uses the same training data and annotations. As shown in Table~\ref{tab:recognition}, ``Ours-SAM-B" surpasses ``ASTER-B" on all benchmarks and the performance gaps are especially large on irregular text benchmarks ($\mathbf{3.7}$ percents on SVTP and $\mathbf{8.3}$ percents on CUTE).

The most related work is \cite{yang2017learning}, which recognizes irregular text with attention mechanisms. It uses a two-class character segmentation for characters detection with character-level annotations to supervise and constructs the ground truth for attention in sequence steps. Different from \cite{yang2017learning}, our model applies a multiple-class character segmentation which can not only produce better representation but also generate text recognition results. Moreover, the spatial attentional module of the standalone recognition model can learn in a weakly-supervised way, without direct ground truth for attention in sequence steps as~\cite{yang2017learning}. For example, ``Ours-SAM-A" and ``Ours-SAM-B" in Table~\ref{tab:recognition} do not use character-level annotations in the training period. The performances in Table~\ref{tab:recognition} also demonstrate the overall superiority of our model against \cite{yang2017learning}.

\subsubsection{Ablation Experiments on the Recognition Model}\label{subsubsec:complement}

\noindent\textbf{Position Embedding}
We conduct experiments to verify the effect of the position embedding by removing it in ``Ours-SAM-without-PE". In Table~\ref{tab:recognition}, ``Ours-SAM-A" achieves better results than ``Ours-SAM-without-PE" among all benchmarks, especially on curved text (CUTE), which demonstrates the advantage of the position embedding. 

\noindent\textbf{Word-level Annotations}
Since SAM requires only word-level annotations, it can use both SynthText and Synth90k in the training period (``Ours-SAM-B"). Compared with the model without using Synth90k (``Ours-SAM-A"), it achieves better results on all benchmarks, as shown in Table~\ref{tab:recognition}. This indicates that using more training data, our model can achieve better results. Considering that some training datasets do not provide character-level annotations, the ability to use word-level annotations is significantly important.

\footnotetext{https://github.com/bgshih/aster}  
\noindent\textbf{Complementarity}
We compare the character segmentation and SAM separately, using the same training data. As shown in Table~\ref{tab:recognition}, ``Ours-segmentation" performs better on IIIT, IC03, and IC13 (most of them are in regular shapes) while ``Ours-SAM-A" is more skilled at ``CUTE" (most of them are in curved shapes). It indicates that they are complementary to each other.
We integrate the character segmentation and SAM into a unified model, which is named as ``Ours-seg-SAM" in Table~\ref{tab:recognition}. Compared with ``Ours-SAM-B", it achieves performance gain on most of the tasks, which further demonstrates the complementarity of the character segmentation and SAM, as the character segmentation module locally predicts the characters while SAM tends to decode the text sequence with global information.

\noindent\textbf{Speed}
The speed of ``Ours-segmentation" is about 50fps with a batch size of 1 while the other variants run at about 20fps. Note that the time costs of the position embedding and the segmentation output in ``Ours-seg-SAM" are ignorable.

\subsection{Summary of the Experiments}
The experimental analysis of the scene text spotting, scene text detection, and scene text recognition among various benchmarks can be concluded as follows:
(1) Compared to the previous text spotters which can only handle horizontal or oriented text, our method can detect and recognize the text of various shapes, including the curved shape.
(2) Our method has a huge superiority on word spotting and end-to-end recognition when no lexicon is given or the given lexicon is of large size, which demonstrates that our method is less reliable to lexicons.
(3) Compared to the conference version, our method not only achieves better performance but also is much more independent of the real-world character-level annotations, which strongly proves the effectiveness of SAM.
(4) Our standalone recognition model surpasses all existing methods on most of the scene text recognition benchmarks and outperforms previous methods by a large margin on the irregular text benchmarks.

\begin{figure}[ht]
\begin{center}
\includegraphics[width=1.0\linewidth]{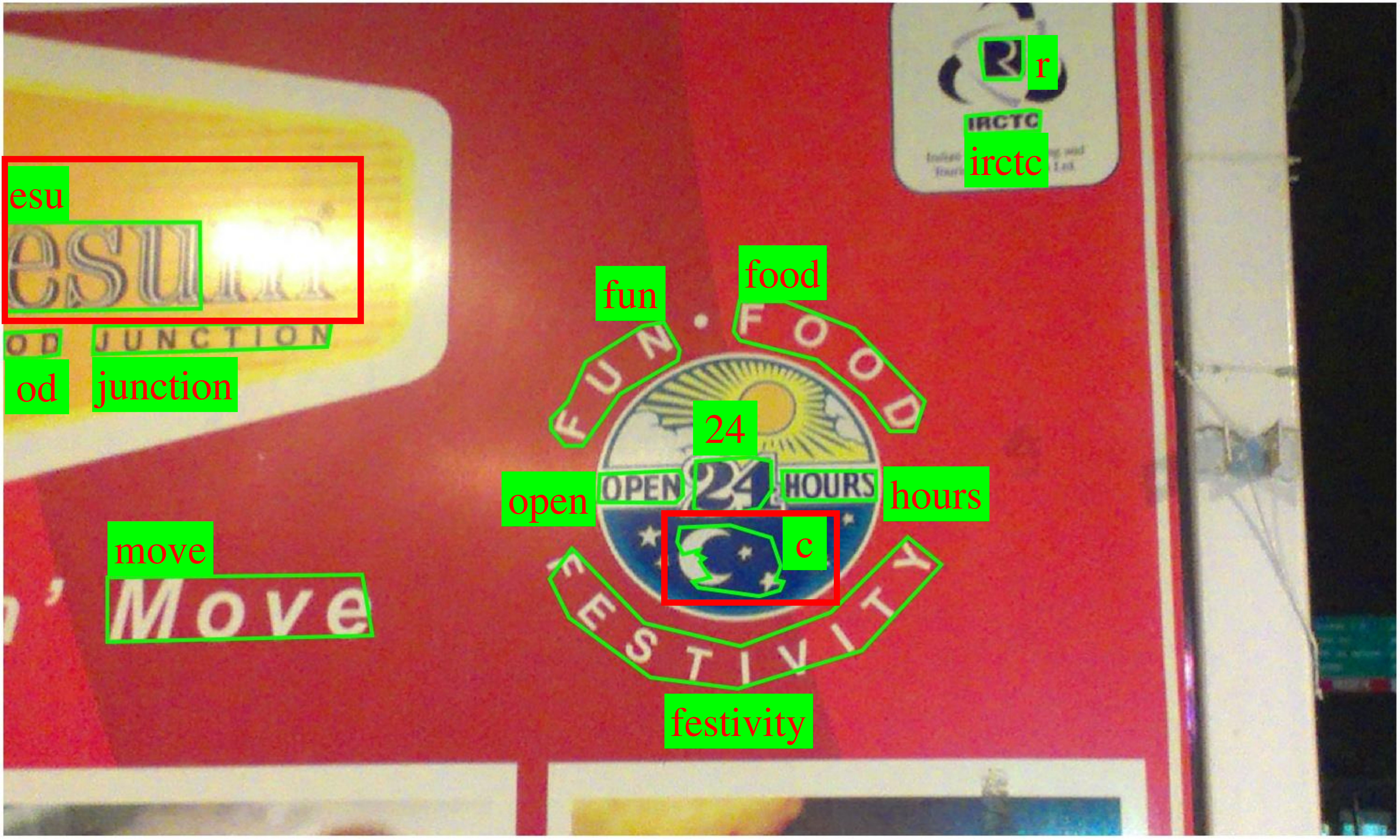}
\end{center}
\caption{Failure cases. Failure cases are in the red boxes for better visualization.}
\label{fig:failure}
\end{figure}

\subsection{Failure Cases}
Some failure cases are visualized in Fig.~\ref{fig:failure}.
One failure case is happened in the detection period due to the extreme illumination. Another failure case is a false positive, where the moon symbol is wrongly-detected and wrongly-recognized as a character.

\section{Conclusion}\label{sec:conclusion}

We have presented Mask TextSpotter, a novel framework of end-to-end text recognition in the wild. Different from the previous text spotters that consider learning-based text recognition as a one-dimensional sequence prediction problem, the proposed method is very easy to train and able to read irregular text, profiting from its two-dimensional representation for both detection and recognition. The state-of-the-art results achieved by Mask TextSpotter in the tasks of scene text detection, scene recognition, and end-to-end text recognition on the standard benchmarks including horizontal text, oriented text, and curved text, validate its generality and effectiveness in reading scene text. 
In the future, we would like to improve the efficiency of Mask TextSpotter, especially exploring to replace the detection stage with more elegant detection method, which is the most time-consuming part of the proposed method.


%

\ifCLASSOPTIONcaptionsoff
  \newpage
\fi



%



{\small
\bibliographystyle{ieee}
\bibliography{references}
}

\end{document}